\documentclass[journal]{IEEEtran}

\usepackage{float}

\usepackage{blindtext}
\usepackage{amsmath, amssymb, amsthm} 
\usepackage[dvipsnames]{xcolor}
\usepackage{hyperref}
\usepackage{flushend}
\usepackage{color, soul} 
\usepackage{comment}
\usepackage{wrapfig}
\usepackage{siunitx}
\usepackage{bm}
\usepackage[utf8]{inputenc} 
\usepackage[T1]{fontenc}    
\usepackage{hyperref}       
\usepackage{url}            
\usepackage{booktabs}       
\usepackage{amsfonts}       
\usepackage{amssymb,amsmath}
\usepackage{nicefrac}       
\usepackage{microtype}      
\usepackage{makeidx}
\usepackage{graphicx} 
\usepackage{subfiles} 
\usepackage{multirow}
\usepackage[]{algorithm2e}
\usepackage[noadjust]{cite}
 
\DeclareMathOperator*{\argmin}{arg\,min}
\DeclareMathOperator*{\argmax}{arg\,max}

\begin{document}
\bstctlcite{IEEEexample:BSTcontrol}

\title{Efficient Hyperdimensional Computing with \\ Modular Composite Representations}


\author{Marco Angioli, Christopher J. Kymn, Antonello Rosato, Amy Loutfi, Mauro Olivieri, and Denis Kleyko
\thanks{
The work of CJK was supported by the Center for the Co-Design of Cognitive Systems (CoCoSys), one of seven centers in JUMP 2.0, a Semiconductor Research Corporation (SRC) program sponsored by DARPA, in addition to the NDSEG Fellowship, Fernström Fellowship, Swartz Foundation, and NSF Grants 2147640 and 2313149.
The work of AL and DK was supported by Knut and Alice Wallenberg Foundation under the Wallenberg Scholars program (Grant No. KAW2023.0327).
DK acknowledges funding from the Swedish Strategic Research Foundation under the Future Research Leaders program (Grant No. FFL24-0111) and the Swedish Research Council under the Starting Grant program (Grant No. 2025-05421).
\textit{Corresponding authors: Marco Angioli and Denis Kleyko}.
}
\thanks{M. Angioli, A. Rosato, and M. Olivieri are with the Department of Information Engineering, Electronics and Telecommunications at Sapienza University of Rome, 00184 Rome, Italy (\mbox{e-mails}: \mbox{\{marco.angioli; antonello.rosato; mauro.olivieri\}@uniroma1.it}).}%
\thanks{C. J. Kymn is with the Redwood Center for Theoretical Neuroscience at the University of California at Berkeley, CA 94720, USA (\mbox{e-mail}: \mbox{cjkymn@berkeley.edu}). }
\thanks{A. Loutfi is with the AI, Robotics and Cybersecurity Center (ARC), and with the Department of Computer Science at Örebro University, 70281 Örebro, Sweden and also with Department of Science and Technology at Linköping University, 58183 Linköping, Sweden (\mbox{e-mail}: \mbox{amy.loutfi@oru.se}).}
\thanks{D. Kleyko is with the AI, Robotics and Cybersecurity Center (ARC), and with the Department of Computer Science at Örebro University, 70281 Örebro, Sweden and also with Intelligent Systems Lab at RISE Research Institutes of Sweden, 16440 Kista, Sweden (\mbox{e-mail}: \mbox{denis.kleyko@oru.se}).}%
}

\maketitle

\begin{abstract}
    The modular composite representation (MCR) is a computing model that represents information with high-dimensional integer vectors using modular arithmetic. Originally proposed as a generalization of the binary spatter code model, it aims to provide higher representational power while remaining a lighter alternative to models requiring high-precision components. However, despite this potential, MCR has received limited attention in the literature. Systematic analyses of its trade-offs and comparisons with other models, such as binary spatter codes, multiply-add-permute, and Fourier holographic reduced representation, are lacking, sustaining the perception that its added complexity outweighs the improved expressivity over simpler models.    
    In this work, we revisit MCR by presenting its first extensive evaluation, demonstrating that it achieves a unique balance of information capacity, classification accuracy, and hardware efficiency. Experiments measuring information capacity demonstrate that MCR outperforms binary and integer vectors while approaching complex-valued representations at a fraction of their memory footprint. Evaluation on a collection of $\mathbf{123}$ classification datasets confirms consistent accuracy gains and shows that MCR can match the performance of binary spatter codes using up to $\mathbf{4.0\times}$ less memory.
    We investigate the hardware realization of MCR by showing that it maps naturally to digital logic and by designing the first dedicated accelerator for it. Evaluations on basic operations and seven selected datasets demonstrate a speedup of up to three orders-of-magnitude and significant energy reductions compared to a software implementation. Furthermore, when matched for accuracy against binary spatter codes, MCR achieves on average $\mathbf{3.08\times}$ faster execution and $\mathbf{2.68\times}$ lower energy consumption. 
    These findings demonstrate that, although MCR requires more sophisticated operations than binary spatter codes, its modular arithmetic and higher per-component precision enable much lower dimensionality of representations. When realized with our dedicated hardware accelerator, this results in a faster, more energy-efficient, and high-precision alternative to existing models.

\end{abstract}

\begin{IEEEkeywords}
Modular Composite Representation, Hyperdimensional Computing, Hardware Acceleration
\end{IEEEkeywords}

\section{Introduction}

    Hyperdimensional computing, also known as vector symbolic architectures (HD/VSA), is a computing paradigm that equips the flexibility of connectionist models with structured transformations of vectors~\cite{Kanerva2009, GaylerJackendoff2003}. At its core, HD/VSA represents information using high-dimensional distributed randomized vectors, known as hypervectors (HVs), which are combined and compared through simple vector arithmetic~\cite{kleyko_review_I, schlegel2022comparison}. This principle endows HD/VSA with several attractive properties, including robustness to noise, energy and computational efficiency, and inherent parallelism, that have made HD/VSA a compelling and promising alternative for deployment in resource-constrained scenarios, digital hardware accelerators, and novel neuromorphic devices~\cite{kleyko_pieee}, for modeling and solving classification~\cite{kleyko_review_II,verges2025classification}, regression~\cite{RegHD, frady2022computing}, clustering~\cite{HDCluster, bandaragoda2019trajectory}, and reinforcement learning tasks~\cite{bees2015imitation, ni2022hdpg, angioli2025hd}. 
    
    Over the years, HD/VSA has been implemented using many different kinds of vector spaces, with data types ranging from binary to real and complex numbers, each striking a different balance between information capacity and computational efficiency~\cite{kleyko_review_I, schlegel2022comparison}.  
    Within this spectrum, the \textit{modular composite representation} (MCR) model, introduced in 2014 and based on modular integer arithmetic, generalizes binary HD/VSA, aiming to increase representational power while remaining a lighter alternative to models requiring floating-point arithmetic. 
    Yet, despite this potential, MCR has received limited attention in the literature, probably because it is perceived neither as efficient as binary HD/VSA nor as expressive as integer- or real-valued models. 

    In this article, we present the first extensive study of MCR, guided by two questions: \emph{when and why is MCR beneficial?} and \emph{what are its execution time and energy consumption when implemented in dedicated hardware?}
    We answer them across three complementary dimensions.
    First, in Section~\ref{sec:res:capacity}, we show that MCR achieves substantially higher \textit{information capacity} than binary and low-precision integer HVs, while approaching the expressiveness of complex-valued models at a fraction of the memory footprint.  
    Second, in Section~\ref{sec:res:acc}, through large-scale experiments on $123$ classification datasets, we demonstrate that \emph{per-component precision matters}: the modular discretized space of MCR consistently outperforms binary and low-precision integer HVs under both equal dimensionality and equal memory footprint settings, and can reduce the HV dimensionality by up to $4\times$ while still surpassing binary models.    
    Finally, in Section~\ref{Section: Hardware Design}, we demonstrate that the modular arithmetic at the core of MCR maps seamlessly to digital hardware and we design the first \textit{dedicated accelerator for MCR}, extending the RISC-V accelerator for binary HD/VSA~\cite{HDCU}. Hardware experiments on the basic operations and seven classification datasets provide details about its execution time, energy efficiency, and scaling behavior. In addition, by directly comparing hardware-accelerated MCR with hardware-accelerated binary models (Section~\ref{section:res:hardware:perf}), we demonstrate that although MCR requires more sophisticated arithmetic, its lower HV dimensionality paired with efficient hardware support translates into a faster, more energy-efficient, more compact, and higher-precision alternative to existing HD/VSA models. 

\section{Background and methods}

\subsection{Hyperdimensional Computing/Vector Symbolic Architectures}
    HD/VSA are a family of neuro-inspired computational models that leverage the mathematics of high-dimensional spaces to represent, combine, and manipulate information~\cite{schlegel2022comparison}. The motivation behind HD/VSA stems from the observation that brains operate with massively parallel circuits built from unreliable components, yet still achieve remarkable robustness and generalization~\cite{Kanerva2009}. Building on this analogy, HD/VSA represents information using HVs where information is \textit{distributed}: each component contributes equally and independently to the overall representation~\cite{kleyko_review_I}. This distributed and holistic representation confers two crucial properties to HD/VSA. First, it provides \emph{robustness to noise and faults}, since even when a large fraction of components are corrupted, the overall representation can still be reliably recognized~\cite{imani2017exploring,rahimi2017high}. Second, it offers \textit{inherent parallelism}, because all HV components can be processed independently, making HD/VSA particularly well-suited for efficient implementations in digital hardware~\cite{RW_Benini, tinyHD, 2021cellular, FixedHD, HDCU} as well as emerging neuromorphic devices~\cite{frady2019robust,9892030,orchard2024efficient}.
    
    Numerous HD/VSA models have been proposed, yet they all support the same basic operations~\cite {schlegel2022comparison,verges2025classification}. At the core of these models lies the fact that fundamental concepts, symbols, or elements can be mapped to random HVs that, thanks to the geometric properties of high-dimensional spaces, are nearly orthogonal, i.e., linearly independent~\cite{Kanerva2009}. These HVs can then be combined and manipulated using a small set of basic operations:
    \begin{itemize}
        \item \emph{Binding} ($\circ$): associates arguments (e.g., key–value pairs) into a new HV that is dissimilar to the arguments.
        \item \emph{Superposition} ($+$): combines multiple HVs into a single composite HV that remains similar to each argument. Since this aggregation can produce components outside the admissible dynamic range,  a subsequent \emph{normalization} step can be applied to project the HV back into its original domain.  
        \item \emph{Permutation} ($\rho$): reorders the components of an HV to produce a dissimilar one, and is used to represent order or role information, such as in sequences.    
    \end{itemize}
    Finally, a \emph{distance} metric ($\delta$) quantifies how close two HVs are in high-dimensional space, serving as the basic mechanism for recognition and retrieval.
    Together, these four operations form the algebra of HD/VSA, enabling the composition of complex information \emph{without increasing dimensionality}.
    
    The implementation of these operations varies depending on the characteristics of the vector space in which HVs are defined~\cite{schlegel2022comparison}.  The components of HVs can be of different \emph{data types}, with each choice giving rise to a distinct HD/VSA model. For instance, the binary spatter codes (BSC) model~\cite{kanerva1995family} operates with binary HVs, while the multiply–add–permute (MAP)~\cite{map} model exists in several variants: MAP-B with binary/bipolar values, MAP-I with integers, and MAP-C with real numbers. Holographic reduced representations (HRR) and Fourier holographic reduced representations (FHRR)~\cite{PlateHolographic2003} employ real and complex numbers, respectively. 
    This variety of models points to the trade-off between information capacity and computational complexity, enabling adaptation to different application domains and hardware platforms. For example, BSC is often favored in resource-constrained scenarios~\cite{Eggimann2021} and in-memory computing~\cite{karunaratne2020memory}, whereas complex-valued FHRR is attractive for neuromorphic hardware~\cite{frady2019robust, orchard2024efficient, 11037669, kymnoscillator}.
    
    \begin{figure}[!t]
        \centering
        \includegraphics[width=0.85\columnwidth]{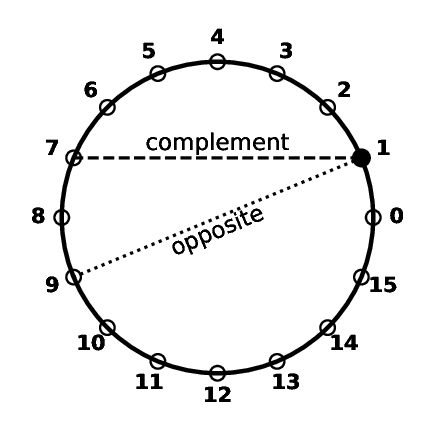}
        \caption{Redrawn illustration from~\cite{MCR} showing the possible discretized values in $\mathbb{Z}_{16}$ for a single component of an MCR HV with $r = 16$. The \emph{complement} of a value is defined as the element whose sum with it equals $r$, while the opposite is its antipode, i.e., the value obtained by adding $r/2$.}
        \label{fig: mcr_discretized_space}
    \end{figure}

\subsection{Modular Composite Representation}\label{Section: MCR}
    MCR is an HD/VSA model introduced in~\cite{MCR} where information is represented with vectors consisting of $D$ integer-valued components. MCR relies on modular arithmetic to define its operations. Each component of an HV lies on a discretized unit circle $\mathbb{Z}_r$, where $r$ is the chosen modulus, so that values range from $0$ to $r-1$ and incrementing $r-1$ wraps around to $0$. 
    Modular vectors form a \textit{homogeneous space} in the sense that for any two vectors $\mathbf{h}$ and $\mathbf{u}$, there exists a third vector $\mathbf{c}$ such that binding $\mathbf{h}$ with $\mathbf{u}$ produces $\mathbf{c}$.
    Fig.~\ref{fig: mcr_discretized_space} illustrates an example of this discretized space with $r=16$, where the complement of a value is defined such that their sum equals $r$, and the opposite is obtained by adding $r/2$.
    
    Within this space, MCR defines binding between two HVs $\mathbf{h}$ and $\mathbf{u}$ as the component-wise modular sum ($c_i=\operatorname{mod}_r({h}_i+{u}_i)$) and unbinding as the component-wise modular subtraction ($c_i=\operatorname{mod}_r({h}_i-{u}_i)$). Similarity is measured by a modular variant of the Manhattan distance that is defined as follows:
    \begin{equation}\label{eq: similarity}
        \delta(\mathbf{h},\mathbf{u}) = \sum_{i=1}^D \min(\operatorname{mod}_r({h}_i-{u}_i), \operatorname{mod}_r({u}_i-{h}_i)).
    \end{equation}
    
    The superposition (also called bundling) of $m$ HVs $\mathbf{h}^{(1)},\dots,\mathbf{h}^{(m)}$ is carried out in two steps. First, each component ${h}^{(j)}_i \in [0,r-1]$ of the $ j$-th HV is projected to an equivalent point on the unit circle (cf. Fig.~\ref{fig: mcr_discretized_space}), and summed in $\mathbb{C}$:
    \begin{equation}\label{eq: bundling}
        v_i = \sum_{j=1}^m \Big(\cos\!\big(\tfrac{2\pi}{r} h^{(j)}_i\big) 
            + \mathrm{i} \sin\!\big(\tfrac{2\pi}{r} h^{(j)}_i\big)\Big).
    \end{equation}
    \noindent
    Next, since the resultant HV $\mathbf{v}$ no longer lies on the discretized unit circle $\mathbb{Z}_r$, its components are mapped back to the nearest elements in $\mathbb{Z}_r$ by finding integers from $[0,r-1]$ with phases that are nearest to the phases of components of the resultant HV:
    \begin{equation}\label{eq: normalization}
        w_i = \operatorname{mod}_r\!\left(
            \left\lfloor 
                \tfrac{r}{2\pi}\,
                \operatorname{mod}_{2\pi}\!\big(\operatorname{atan2}(\Im(v_i),\,\Re(v_i))\big)
            \right\rceil
        \right).
    \end{equation}

    If one component happens to be zero (e.g., when an element is superimposed with its inverse), its undefined phase is resolved by assigning it to the integer closest to the arithmetic mean of the original arguments in the integer domain:
    \begin{equation}\label{eq: bundling_normalization}
        w_i \;=\; \operatorname{mod}_r\!\left(
            \left\lfloor 
                \tfrac{1}{m}\sum_{j=1}^{m} h^{(j)}_i
            \right\rceil
        \right).
    \end{equation}

    \noindent
    Notably, each discretization step (hereafter referred to as \textit{normalization}) results in information loss. Therefore, when superimposing multiple HVs, the normalization in Eq.~(\ref{eq: normalization}) should be deferred until the final step~\cite{MCR}.
    
    MCR can be seen as a middle-ground between the properties and advantages of BSC and FHRR.
    It was originally proposed as a generalization of BSC, inheriting its low memory requirements~\cite{MCR}. In fact, when $r=2$, MCR is equivalent to BSC. For $r>2$, however, its operations generalize the modular arithmetic of BSC beyond $\mathbb{Z}_2$.
    At the same time, under the interpretation of integers in $[0,r-1]$ as phasors on the unit circle (the $r$th roots of unity), the binding and superposition operations of MCR implement a discretized FHRR, offering comparable expressiveness while being more computationally efficient and easier to implement with LUTs (as we report in Section~\ref{Section: hardware_friendly}). In this sense, MCR can be seen as a model covering the whole spectrum between BSC and FHRR, balancing expressiveness and efficient implementation. 
    In line with this discretized-phase view, Yu et al.\ introduced the Cyclic Group Representation (CGR) model~\cite{yu2022understanding}, which, similar to MCR implements the binding operation as the component-wise modular sum. Analysis in~\cite{yu2022understanding} shows that certain similarity functions cannot be expressed by binary HVs but can be realized within CGR; accordingly, MCR with $r>2$ also possesses this broader expressivity. Furthermore, modules of MCR/CGR can be combined to form residue number systems~\cite{kymn2025computing}, which has promising applications to unconventional computing~\cite{omondi2007residue} and neuroscience~\cite{fiete2008grid,kymn2024binding}. 
    
    No prior work has systematically investigated MCR. Beyond being briefly mentioned in a survey~\cite{kleyko_review_I} and implemented in software~\cite{torchhd}, its performance on real datasets and potential for hardware implementation have not been thoroughly evaluated. 
    The existing literature has predominantly focused on the two ``extreme'' points of the design spectrum, BSC and FHRR, while largely overlooking integer-valued alternatives. Perhaps this was because the added computational cost of such designs seemed unjustified relative to their potential gains over binary HVs. Even the original study~\cite{MCR} conjectured that the simpler operations of BSC might be preferable in resource-constrained scenarios with strict timing requirements. In this study, we revisit that conjecture by systematically analyzing the algorithmic performance and hardware complexity of MCR, demonstrating that it offers a favorable trade-off between HV dimensionality and component dynamic range compared to many existing HD/VSA models.

    \subsection{Capacity of hypervectors} 
    \label{sec:met:capacity}
    
    In HD/VSA, the \textit{information capacity} is measured as the maximum amount of information (bits) that can be decoded from an HV representing a data structure. 
    
    We adopt the setup from~\cite{FradyCapacity2018, hersche2021near,kleyko2023efficient} to evaluate the information capacity of MCR and compare it with the well-known BSC, MAP-I, and FHRR models. Given a codebook matrix $\Phi$ of $d$ symbols, where each column $\Phi_{j}$ is a fixed $D$-dimensional HV representing symbol $j$, we uniformly sample a sequence $\mathbf{s}=(s_1,\dots,s_m)$ of length $m$ and construct its composite HV as follows:
    \begin{equation}\label{eq:capacity_composite_representation}
        \phi(\mathbf{s}) = \sum_{j=1}^{m}\rho^{\,m-j}(\Phi_{s_j}),
    \end{equation}
    \noindent
    where $\rho(\cdot)$ is a permutation operation representing the position of each symbol within the sequence.

    The decoding task then consists of reconstructing the original sequence $\mathbf{s}$ from $\phi(\mathbf{s})$, producing an estimate $\hat{\mathbf{s}}$ as close as possible to $\mathbf{s}$. This can be done using various decoding techniques (see, e.g.,~\cite{hersche2021near,kleyko2023efficient}). 
    In this study, we use the simplest one -- \textit{Codebook decoding}, where each symbol in the codebook is compared to the permuted composite HV $\rho^{-(m-j)}(\phi(\mathbf{s}))$ using an appropriate distance metric, $\delta$, and the symbol with the lowest distance is selected as the estimate $\hat{s}_j$, as in Eq.~\eqref{eq:capacity_codebook_decoding}. This step is repeated for all positions $j \in [1,m]$ to decode the complete sequence $\hat{\mathbf{s}}$.
    \begin{equation}\label{eq:capacity_codebook_decoding}
        \hat{s}_j = \argmin_{c \in \{1,\ldots,d\}}\delta( \Phi_c , \rho^{-(m-j)}(\phi(\mathbf{s}))).
    \end{equation}
    
    Following the setup in~\cite{hersche2021near,kleyko2023efficient}, the decoding performance and, thus, the capacity of HVs, are evaluated in terms of two performance metrics. The first metric is the \textit{decoding accuracy}, defined in Eq.~\eqref{eq: capacity_accuracy} as the proportion of correctly decoded symbols, averaged over $g$ trials.
    \begin{equation}\label{eq: capacity_accuracy}
        a= \frac{1}{gm}\sum_{k=1}^{g}\sum_{j=1}^{m}\mathbb{I}[s^{(k)}_j = \hat{s}^{(k)}_j].
    \end{equation}
    
    The second metric is the \textit{information rate}. For a given decoding accuracy $a$ and codebook size $d$, the amount of information decoded per symbol can be computed as follows:
    \begin{equation}\label{eq: capacity_I_symb}
        I_{\texttt{symb}}(a,d) = a\log_2(da) + (1-a)\log_2\!\left(\frac{d}{d-1}(1-a)\right).
    \end{equation}
    The \textit{total information} decoded from a sequence of length $m$ can then be calculated as:
    \begin{equation}\label{eq: capacity_I_tot}
        I_{\texttt{tot}} = m I_{\texttt{symb}}(a,d).
    \end{equation}
    Normalizing with respect to the dimensionality $D$ yields the \textit{information rate per component} ($I_{\texttt{dim}}$):
    \begin{equation}\label{eq: capacity_I_dim}
        I_{\texttt{dim}} = \frac{I_{\texttt{tot}}}{D}.
    \end{equation}
     Furthermore, we also report the \textit{information rate per storage bit} ($I_{\texttt{bit}}$), Eq.~(\ref{eq: capacity_I_bit}), where $b$ denotes the number of bits required per HV component. This metric accounts for both accuracy and memory footprint and is, therefore, suitable for comparing models with components requiring different precision, such as binary, integer-, real- or complex-valued.
     \begin{equation}\label{eq: capacity_I_bit}
      I_{\texttt{bit}} = I_{\texttt{tot}}/(D b).
     \end{equation}

    \subsection{Classification with HVs}
    \label{Section:methods:Accuracy}

    In this study, we compare the classification performance of different HD/VSA models across a collection of classification datasets with tabular data. Therefore, we use transformation for key–value pairs to represent feature vectors as HVs. For tabular data, this transformation has been shown to achieve the best trade-off between execution time and accuracy in this setting~\cite{kleyko2021density, verges2025classification}. Given a $d$-dimensional feature vector, $\mathbf{x} = [x_1, x_2, \ldots, x_d]$, each feature $j$ is assigned a random HV $\mathbf{r}^{(j)}$ acting as a key. This HV is then bound to a value vector ($\psi(x_j)$) that is obtained via the quantization of the corresponding feature value $x_j$ and mapping it into the high-dimensional space using the thermometer code~\cite{penz1987closeness, Rachkovskiy_2005}. 
    In MCR, the thermometer code is realized by assigning $0$ and $r/2$ as the two extreme values.
    Such mapping ensures that similar feature values are transformed into similar HVs. The outcome of the bindings is $d$ key-value pairs, which are then superimposed, resulting in the composite HV \(\phi(\mathbf{x})\) representing the feature vector:
    \begin{equation}
    \label{eq:class:encod}
        \phi(\mathbf{x}) = \sum_{j=1}^d \mathbf{r}^{(j)} \circ \psi(x_j).
    \end{equation}
    
    To train the classifier, we adopt a prototype-based Learning Vector Quantization (LVQ) approach~\cite{nova2014review, DiaoGLVQHD2021}. During the first epoch, prototypes are initialized using a simple centroid classifier~\cite{kleyko_2015, rahimi2016robust, industrial2018} where all HVs belonging to the same class are aggregated into a class prototype:
    \begin{equation} \label{eq: centroid_classifier}
        \mathbf{p}_c = \sum_{k: y^{(k)}=c} \phi(\mathbf{x}^{(k)}).
    \end{equation} 
    From the second epoch onward, LVQ2.1~\cite{kohonen1990improved, nova2014review} is used according to Eq.~\eqref{eq: LVQ2.1}, where two prototypes are updated per training sample: the correct prototype $\mathbf{p}_+$ and the closest incorrect prototype $\mathbf{p}_-$. For a sample $\phi(\mathbf{x})$ with label $c$, distance is computed using the appropriate model-specific metric: Hamming for BSC, cosine for MAP, modular Manhattan for MCR, Eq.~(\ref{eq: similarity}). 
    \begin{equation}\label{eq: LVQ2.1}
    \begin{cases}
    \mathbf{p}_+ \leftarrow \mathbf{p}_+ + \epsilon(\phi(\mathbf{x})-\mathbf{p}_+) \\
    \mathbf{p}_- \leftarrow \mathbf{p}_- - \epsilon(\phi(\mathbf{x})-\mathbf{p}_{-}).
    \end{cases}
    \end{equation}
    Here $\epsilon$ is the learning rate. The learning rule is triggered only if $\phi(\mathbf{x})$ lies inside the LVQ2.1 window:
    \begin{equation}\label{eq: LVQ_window}
    \min\!\left(\frac{\delta(\phi(\mathbf{x}),\mathbf{p}_-)}{\delta(\phi(\mathbf{x}),\mathbf{p}_+)}, \frac{\delta(\phi(\mathbf{x}),\mathbf{p}_+)}{\delta(\phi(\mathbf{x}),\mathbf{p}_-)}\right) > s,
    \quad s=\frac{1-\omega}{1+\omega},
    \end{equation}
    \noindent
    where in the conducted experiments $\omega$ is set to $0.1$ following~\cite{nova2014review}.
    To prevent the uncontrolled growth of prototypes' norms during iterative updates and ensure the correct operation of LVQ2.1, L2-norms of all class prototypes are reset to one after each training epoch.
    
    To minimize the loss of precision, we maintain high intermediate precision (floating-point) during training and normalize class prototypes only after the end of training. For BSC, this corresponds to the sign function. For MAP-I with low precision, we first rescale the distribution of prototypes' values to match the dynamic range of the targeted $b$-bit interval $[-2^{b-1},2^{b-1}-1]$, and then apply uniform quantization. For MAP-C32 ($32$ bits per component), no normalization is applied, maintaining prototypes in floating-point precision. For MCR, prototype updates are accumulated in the complex domain, Eq.~(\ref{eq: bundling}), and then discretized back to $\mathbb{Z}_r$ as described by Eq.~(\ref{eq: normalization}).
    
    During inference, each test sample is represented following Eq.~(\ref{eq:class:encod}) and classified using the class label of the nearest prototype as follows: 
    \begin{equation} \label{eq: nearest_prototype}
        \hat{y}(\phi(\mathbf{x})) = \arg\min_{c}\;\delta(\phi(\mathbf{x}),\mathbf{p}_c). 
    \end{equation}

    \begin{figure*}
        \centering
        \includegraphics[width=2\columnwidth]{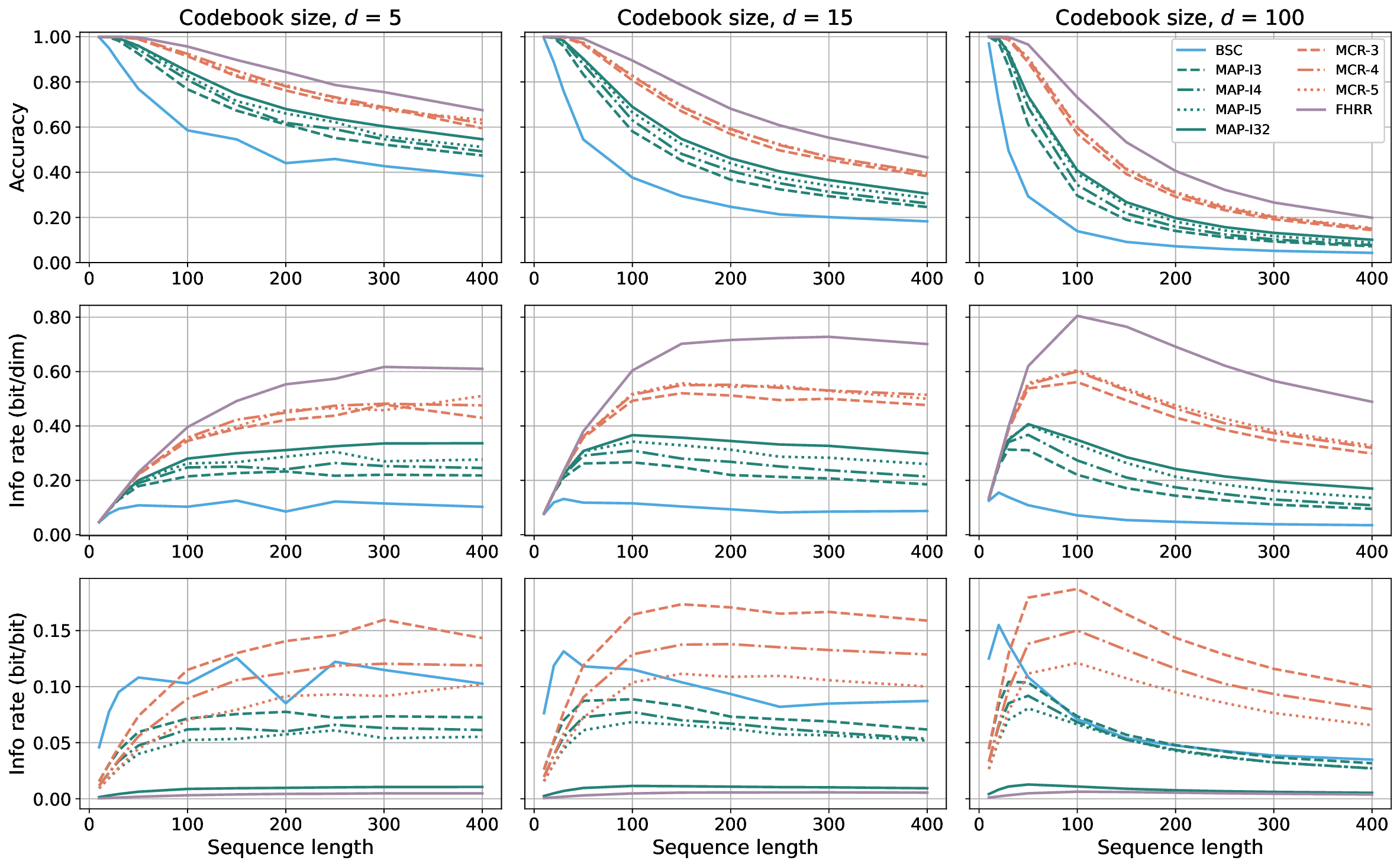}
        \caption{Decoding performance and information capacity analysis for MCR compared to BSC, MAP-I, and FHRR. Results are shown for three different codebook sizes ($d=5,15,100$), using HVs of size $D=500$ and sequence lengths ranging from $m=10$ to $m=400$, averaged over 20 independent codebooks with 50 test sequences per length. MCR variants achieve higher capacity than BSC and MAP-I variants and approach the performance of FHRR while maintaining much greater memory efficiency.}
        \label{fig: capacity_results}
    \end{figure*}

    \subsection{Hyperdimensional Coprocessor Unit}\label{Section: HDCU}
    To demonstrate the practical feasibility of MCR arithmetic in digital hardware, in this study we extend the Hyperdimensional Coprocessor Unit (HDCU) presented in~\cite{HDCU} to support the complete set of MCR operations. 
    The HDCU is a \textit{configurable, general-purpose, open-source} accelerator, originally designed for BSC and tightly coupled with the Klessydra-T03 RISC-V core~\cite{Klessydra-T13}, operating as a coprocessor for the core. Its architecture is based on three guiding principles:  
    \begin{itemize}
        \item \textit{General-purpose HDC acceleration}: instead of hardwiring fixed learning pipelines, the HDCU accelerates the core HD/VSA operations:  binding, superposition, permutation, and distance via dedicated functional units.  
        \item \textit{Software programmability}: a custom RISC-V Instruction Set Extension (ISE) for HD/VSA enables the control of the accelerator via intrinsic functions fully integrated into the GNU Compiler Collection (GCC) toolchain. This allows the same hardware to be programmed for accelerating diverse HD/VSA tasks such as classification, regression, or reinforcement learning. 
        \item \textit{Hardware configurability}: extensive synthesis-time parameters (e.g., degree of parallelism, supported operations, local memory size) allow designers to trade off performance against resource usage, tailoring the accelerator to the requirements of the target platform.  
    \end{itemize}

    Further details on the original HDCU internals and interface can be found in~\cite{HDCU}. 

    \section{Results}

    \subsection{Capacity analysis} 
    \label{sec:res:capacity}

    Following the experiment design introduced in Section~\ref{sec:met:capacity}, we perform experiments to measure the information capacity of MCR and several other well-known HD/VSA models: BSC, MAP-I, and FHRR.

    Fig.~\ref{fig: capacity_results} reports the obtained results for three different codebook sizes ($d \in \{5,15,100\}$), using HVs of size $D=500$ and sequence lengths ranging from $m=10$ to $m=400$, with analysis performed following the methods reported in~\cite{hersche2021near,kleyko2023efficient}. We compare MCR with different quantizations ($r \in \{4,8,16\}$, corresponding to 2, 3, and 4 storage bits per component, respectively), BSC (1 bit per component), MAP-I (2, 3, 4, and 32 bits per component), and FHRR (128 bits per component, with 64 bits each for the real and imaginary parts). For MCR, MAP-I, and BSC, to minimize information loss, the composite HV $\phi(\mathbf{s})$ was first constructed using the full-precision accumulation, and only afterwards normalized back to the target precision. The results are averaged over $20$ independent random codebooks, with $50$ randomly generated sequences for each considered sequence length $m$.
    
    As expected, the results demonstrate that MCR drastically outperforms BSC as the sequence length increases, with an average improvement in decoding accuracy of 25.5\% across sequence lengths from 10 to 400 and the three codebook sizes. More impressively, each MCR variant outperforms the corresponding MAP-I variant with matching precision, yielding average gains of 12.01\% for 3 bits, 10.80\% for 4 bits, and 8.88\% for 5 bits. This positions \textit{the modular space of MCR as a superior choice} over MAP-I quantization for memory-constrained environments. Surprisingly, MCR even performs significantly better than the unconstrained MAP model with 32 bits per component, with an average accuracy gain of 7.63\%. Although this may appear counterintuitive, the difference in performance comes from the different properties of superposition adopted by the two models. While MAP relies on a simple integer sum (in effect, only a real part), MCR interprets integers as discretized phasors on the unit circle, and then performs vector addition in $\mathbb{C}$. 
    The greater expressivity in the complex plane preserves more information during superposition and explains why MCR achieves higher capacity. This fact is also consistent with the strong performance of FHRR decoding, which operates in the same manner except for quantization. FHRR has slightly higher decoding accuracy than MCR, although at the cost of requiring many more bits, as we will see. 
    Notably, increasing $r$ yields diminishing returns in MCR: although higher $r$ reduces phase quantization error, the normalization step discards magnitude, preventing MCR from fully converging to FHRR.  
    
    When the results are analyzed in terms of information per bit, MCR clearly outperforms all other HD/VSA models. Despite its high capacity, FHRR has the lowest $I_{\texttt{bit}}$ due to the 128 bits required per component. For the same reason, MAP-I32 is the second lowest. MAP-I variants with low precision improve memory efficiency by using only three to five bits, but they still underperform compared to 1-bit BSC. 
    In contrast, MCR combines high capacity with low precision, consistently achieving the \textit{highest information-per-bit rate} across all settings.

    \begin{figure}[t]
        \centering
        \includegraphics[width=0.8\columnwidth]{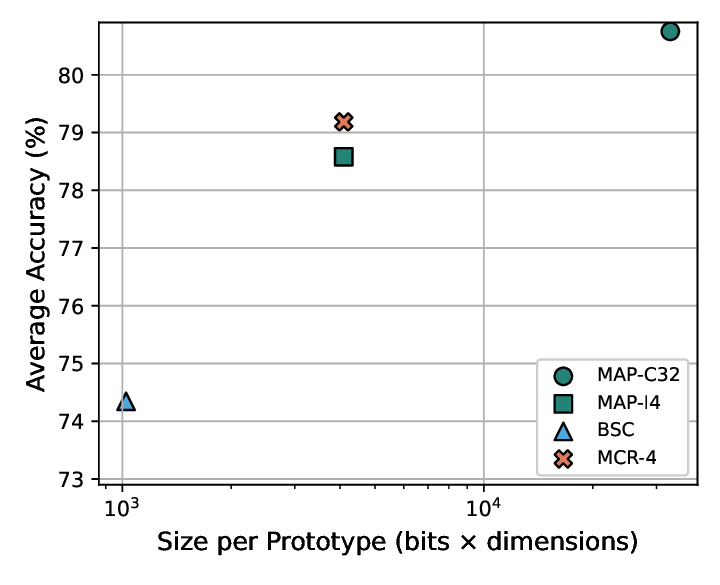}
        \caption{Average accuracy across $123$ classification datasets for 1-bit BSC, MAP-I4, MCR-4 and MAP-C32, using HVs of size $D=1024$. MCR outperforms BSC and MAP-I4 and approaches MAP-C32.}
        \label{fig:torchhd_accuracy_fixed}
    \end{figure}

    \subsection{Classification performance}
    \label{sec:res:acc}

    To evaluate the performance of MCR on real datasets, we carried out a large-scale study on a collection of $123$ classification datasets from~\cite{verges2025classification}. This collection is based on a popular collection from~\cite{HundredsClassifiers2014} representing a subset of 121 datasets from the UCI Machine Learning repository~\cite{Dua2019}. 
    We compare MCR-4 to BSC (1 bit), MAP-I4, and MAP-C32. 
    The objective of this analysis is to understand whether the higher per-component precision and higher capacity of MCR translate into tangible improvements in classification accuracy when applied to diverse data, and whether it provides a better trade-off under strict memory constraints.

    For our experiments, we used an HV dimensionality $D=1024$, key–value transformation with $1024$ quantization levels generated using the thermometer code, $10$ training epochs, and LVQ2.1 with the default parameters: $\omega=0.1$ and $\epsilon=0.01$. We averaged the results over $20$ independent runs per dataset.  
    
\begin{figure}[t]
        \centering
        \includegraphics[width=0.9\columnwidth]{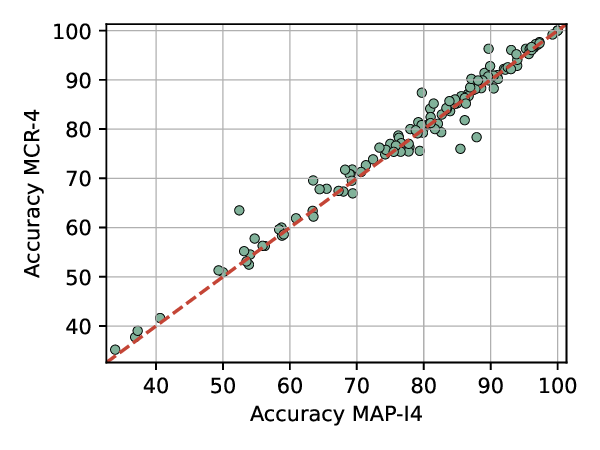}
        \caption{Scatter plot comparing classification accuracy of MAP-I4 ($x$-axis) against MCR-4 ($y$-axis) across $123$ classification datasets. Each point corresponds to one dataset. The dashed diagonal represents equal performance between the two models: the points above the diagonal indicate datasets where MCR-4 outperforms MAP-I4, while the points below indicate the opposite.}
        \label{fig:mcr_vs_mapclip_scatter}
    \end{figure}

    Fig.~\ref{fig:torchhd_accuracy_fixed} reports the results for MCR-4 compared to the baseline models, plotting the average accuracy versus the memory footprint per class prototype (bits\,$\times$\,dimension). The results confirm that MCR-4, with its higher per-component precision, ($D=1024$, $b=4$) outperforms BSC ($D=1024$, $b=1$), achieving an average gain of $+4.84\%$. More importantly, MCR-4 also outperforms MAP-I4 ($D=1024$, $b=4$) under the same memory footprint with an average gain of $+0.61\%$ and approaches the accuracy of full-precision MAP-C32 (-1.57\%) while requiring only a quarter of its memory footprint.
    A complementary view of these results is shown in Fig.~\ref{fig:mcr_vs_mapclip_scatter}, which compares the relative accuracy of MCR-4 and MAP-I4 on each dataset. Most points lie above the diagonal, confirming that MCR provides systematic improvements.   
    
\begin{figure}[t]
        \centering
        \includegraphics[width=0.8\columnwidth]{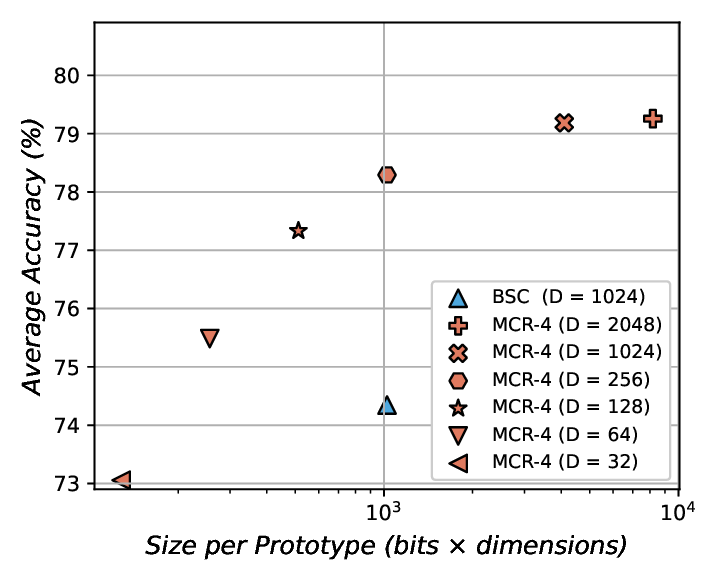}
        \caption{Average classification accuracy across $123$ classification datasets for MCR-4 with different $D$ compared to BSC ($D=1024$, $b=1$). With the same memory footprint, MCR-4 ($D=256$, $b=4$) outperforms BSC by +3.94\% on average. Even at $D=64$, requiring only a quarter of BSC's memory, MCR-4 still achieves a $+1.14\%$ improvement, highlighting its superior accuracy–memory trade-off.}
        \label{fig:torchhd_accuracy_scaling}
    \end{figure}

    To further investigate the accuracy–memory trade-off, Fig.~\ref{fig:torchhd_accuracy_scaling} presents the results for MCR-4 with a reduced number of components in HVs. When $D$ is reduced to $256$, MCR-4 has the same memory footprint as BSC ($D=1024$, $b=1$), MCR-4 still outperforms the binary model with an average improvement of {+3.94\%}. Even more impressively, MCR-4 remains robust when dimensionality is further decreased: with only $64$ components, MCR-4 still surpasses BSC ($+1.14\%$) while requiring $75\%$ less memory.     
    
    These results clearly demonstrate that \textit{per-component precision matters}: the few-bit quantization defined by MCR enables superior accuracy–memory efficiency compared to simple binarization or restricting the integer range with the clipping function. It also supports a substantial reduction in dimensionality while maintaining competitive classification performance.

\subsection{Hardware design for MCR}
\label{Section: Hardware Design}
    The previous sections demonstrated that MCR achieves a superior trade-off between memory footprint, classification accuracy, and information capacity compared to other well-known HD/VSA models.
    However, a natural concern is whether these improvements come at the expense of additional computational costs. Indeed, the original MCR study~\cite{MCR} conjectured that BSC, with its simpler operations, could be preferable for resource-constrained scenarios.
    This raises an important question: \textit{can MCR-specific hardware design solve the impact of the additional computational cost of MCR on execution time and energy consumption?}
    In this section, we provide a solution by moving from software simulations to hardware design, showing that the arithmetic of MCR naturally maps onto digital circuits and presenting the first dedicated hardware accelerator for this HD/VSA model.

    \subsubsection{Hardware-friendly MCR arithmetic}\label{Section: hardware_friendly} 
    There are two aspects that might initially be perceived as computationally expensive in MCR: the use of modular reductions and the reliance on trigonometric functions to perform normalization.
    \paragraph{Modular reductions}   
    As described in Section~\ref{Section: MCR}, the binding, unbinding, and distance operations in MCR require applying a modular reduction ($\operatorname{mod}_r$) for every HV component to remain in $\mathbb{Z}_r$. 
    This frequent operation might seem costly, since modulo is normally computed in hardware by a divider unit: division is neither associative nor distributive, making parallelization difficult and leading to high area and power costs even with optimized dividers~\cite{vlnpd}. However, when implemented in digital hardware with $r$ chosen as a power of 2, modular reduction actually comes for free: if each component is stored using $b=\lceil \log_2(r)\rceil$ bits, the modulo operation is automatically performed by binary overflow. For example, with $r=16$, storing each component with 4 bits ensures that $14+5=19$ overflows to $3$, which is exactly $\operatorname{mod}_{16}(19)$. 
    As a result, \textit{binding is reduced to integer addition, unbinding to integer subtraction, and distance computation to two subtractions followed by a $\min$ operation}. 
    These operations are lightweight, fully parallelizable, and incur no additional cost beyond the standard integer arithmetic.

    \paragraph{Trigonometric mappings}  
    Another potentially costly step is the involvement of trigonometric functions. 
    First, during superposition, each component $h^{(j)}_i \in [0,r-1]$ must be mapped to a unit vector in $\mathbb{R}^2$ using $\cos$ and $\sin$ functions. 
    In software, this requires repeated evaluations of trigonometric functions, which are computationally expensive. 
    In hardware, however, the discretized modular space makes this step efficient: all $\cos$ and $\sin$ values for all values in $\mathbb{Z}_r$ can be pre-computed and stored in compact fixed-point LUTs, so that each mapping reduces to a single memory lookup followed by a fixed-point accumulation. 
    This completely removes the need for trigonometric evaluation during runtime.
    
    Second, during the normalization, the accumulated HV in $\mathbb{C}$ must be projected back to the nearest integer in $\mathbb{Z}_r$ using $\operatorname{atan2}$, Eq.~\eqref{eq: normalization}. A direct hardware realization of this step is costly, as it requires fixed-point division and nonlinear functions. 
    Division, as already discussed, is slow and difficult to parallelize. 
    Similarly, approximating $\operatorname{atan2}$ by computing the ratio $\Im(v_i)/\Re(v_i)$ and mapping it through a LUT demands a relatively large memory, since the ratio may assume many values depending on fixed-point precision, creating an unfavorable trade-off between accuracy, area, and power consumption.
    To overcome these limitations in the hardware realization, two approaches can be adopted.  
    The first one is the common CORDIC algorithm, which computes $\operatorname{atan2}(\Im(v_i),\Re(v_i))$ iteratively using only shift-and-add operations and a small LUT of arctangent constants.  
    The second one, proposed in this work, is a winner-take-all (WTA)-based approach: 
    \begin{equation}\label{eq: WTA}
    w_i = \argmax_{k\in[0,r-1]} \; \big(\Re(v_i) \cos\!\left(\tfrac{2\pi}{r}k\right) + \Im(v_i) \sin\!\left(\tfrac{2\pi}{r}k\right)\big).
    \end{equation}
    The key idea is to compare $v_i$ against all $r$ discrete directions $(\cos \tfrac{2\pi}{r}k, \sin \tfrac{2\pi}{r}k)$ retrieved from the LUTs and select the one with the highest inner product.
    Importantly, since the quadrant of $v_i$ can be identified directly from the sign bits of $\Re(v_i)$ and $\Im(v_i)$, the search must cover only $r/4+1$ values in the corresponding quadrant, rather than all $r$ directions.
    This reduces the normalization to $r/4+1$ inner product computations (two multiplications and one addition each), followed by the search for an argument of the maximum. 
    In discretized spaces, this normalization procedure is highly efficient and avoids division. Unlike CORDIC, it is fully parallelizable across all candidate directions while being more compact and not requiring additional LUTs.
        
    \subsubsection{MCR-HDCU}
    Building on the foundation of HDCU (Section \ref{Section: HDCU}), we design MCR-HDCU by preserving its integration strategy while replacing the binary datapath with one that is tailored for modular operations, achieving software-controlled and efficient acceleration while retaining full flexibility and programmability. To the best of our knowledge, this represents the first dedicated hardware accelerator for MCR.     
    
    \begin{figure}[!t]
        \centering
        \includegraphics[width=0.9\columnwidth]{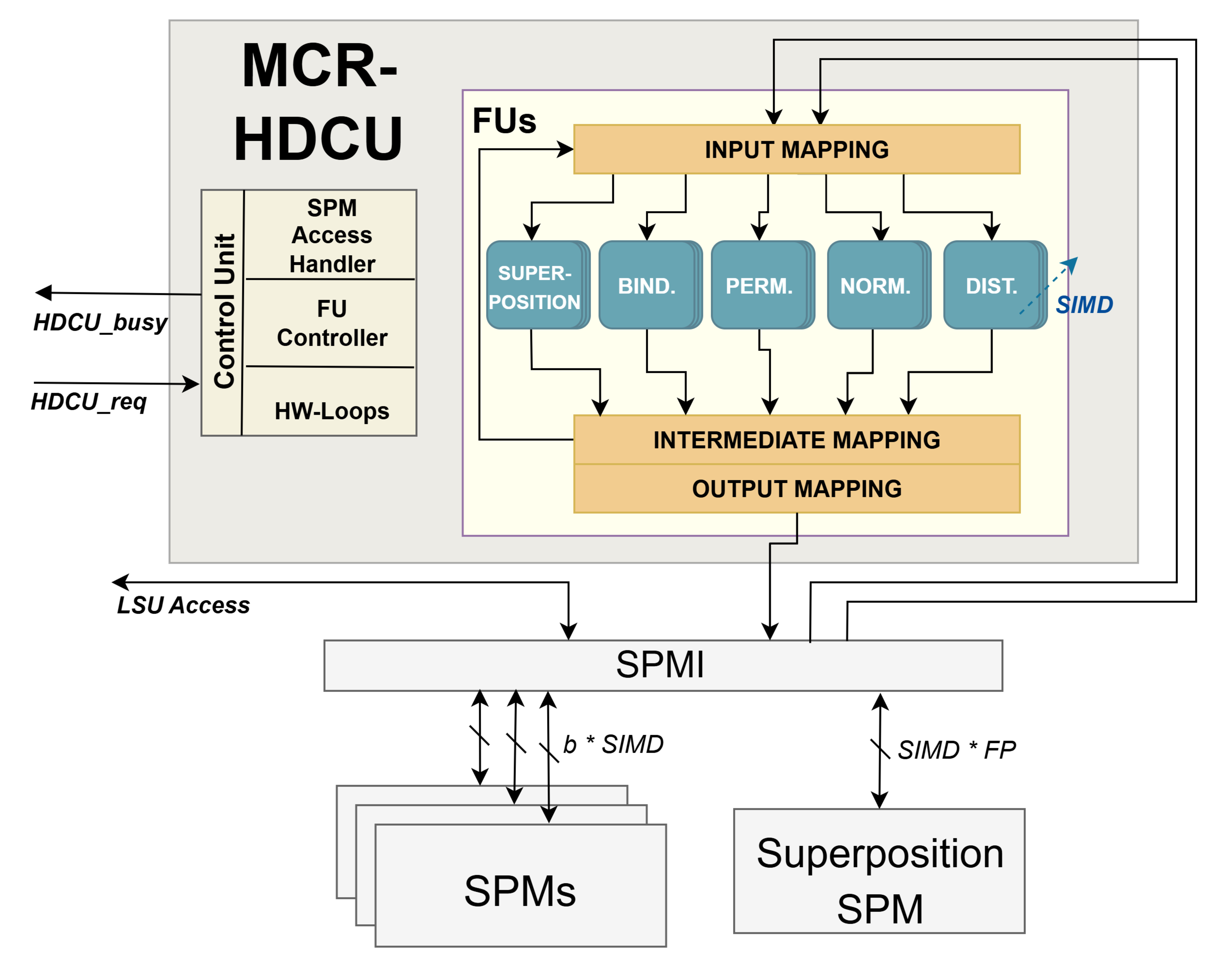}
        \caption{High-level schematic of the MCR-HDCU microarchitecture. The design integrates dedicated functional units for each arithmetic operation in MCR, running in parallel over $\texttt{SIMD}$ lanes.  
        Specialized Scratchpad Memories ensure low-latency, high-bandwidth access. The hardware parallelism and memory configuration (number and size of Scratchpad Memories) can be configured at synthesis time.}
        \label{fig: MCR_HDCU}
    \end{figure}
    
    Fig.~\ref{fig: MCR_HDCU} depicts a high-level view of the MCR-HDCU microarchitecture. The accelerator features dedicated pipelines and optimized functional units (FUs) for each MCR arithmetic operation. 
    As in the original HDCU, each FU can be selectively enabled or disabled at synthesis time, and its degree of hardware parallelism can be configured through the \texttt{SIMD} (Single Instruction Multiple Data) parameter. 
    The design of each functional unit in MCR-HDCU incorporates the optimized arithmetic strategies introduced in Section~\ref{Section: hardware_friendly}, ensuring efficient implementation of the modular operations of MCR.
    
    Each HV component is represented using $b = \lceil \log_2(r)\rceil$ bits, where $r$ is a synthesis-time parameter. With this representation, modular wrap-around is naturally handled by binary overflow, eliminating explicit modulo operations and directly wrapping in the $\mathbb{Z}_r$ ring, as discussed in Section~\ref{Section: hardware_friendly}.
    
    \begin{table}[!t]
        \centering
        \caption{MCR-HDCU Configuration Parameters}
        \label{tab: HDCU_Configuration}
        \resizebox{0.9\columnwidth}{!}{\begin{tabular}{c|c}
        \textbf{Parameter} & \textbf{Configuration Time} \\
        \midrule
        SPM Size & Synthesis \\
        SPM Number & Synthesis \\
        Hardware Parallelism (\texttt{SIMD}) & Synthesis \\
        Functional Unit Enable/Disable & Synthesis \\
        MCR Modulo ($r$) & Synthesis \\
        Fixed-point Precision (\texttt{FP}) & Synthesis \\
        \texttt{HVDIM} & Runtime \\
        \texttt{HVCLASS} & Runtime \\
        \end{tabular}}
    \end{table}
        
    To sustain parallel HV operations, the accelerator integrates local dedicated Scratchpad Memories (SPMs) that provide low-latency, high-bandwidth access. 
    The number and size of these SPMs are configurable at synthesis time. 
    By default, four SPMs are instantiated: three with bandwidth $b \texttt{SIMD}$, sustaining the transfer of \texttt{SIMD} components per cycle, and one with higher bandwidth $ \texttt{FP} \cdot \texttt{SIMD}$ dedicated to the Superposition Unit, which requires streaming \texttt{SIMD} fixed-point components per cycle for Cartesian accumulation, where \texttt{FP} denotes the fixed-point precision.
    When the target learning task is known in advance, the SPMs dimension can be set precisely to the required number of HVs; for more general-purpose deployments, larger SPMs may be synthesized to preserve flexibility. 
    Data exchange between the host core and the SPMs is supported by dedicated \texttt{hvmemld} and \texttt{hvmemstr} instructions~\cite{HDCU}, though in typical workloads, the accelerator operates autonomously on locally stored HVs.  Table~\ref{tab: HDCU_Configuration} summarizes the parameters that can be tuned at the synthesis time and at runtime in MCR-HDCU.
    
    Next, we describe the implementation of each arithmetic operation in the MCR-HDCU accelerator.
    
    \paragraph{Binding}  
    By representing HV components using $b$ bits, binding is implemented as a simple modular addition, with the \texttt{SIMD} parameter defining how many adders operate in parallel. The same holds for unbinding, which is a modular subtraction. Since additions/subtractions are fully parallelized over the \texttt{SIMD} lanes, the total latency in clock cycle ($L$) is  
    \[
    L_{\texttt{bind}} = L_{\texttt{unbind}} = \frac{\texttt{HVDIM}}{\texttt{SIMD}} \;\; \text{cycles}
    \]  
    
\begin{figure}[!t]
        \centering
        \includegraphics[width=0.8\columnwidth]{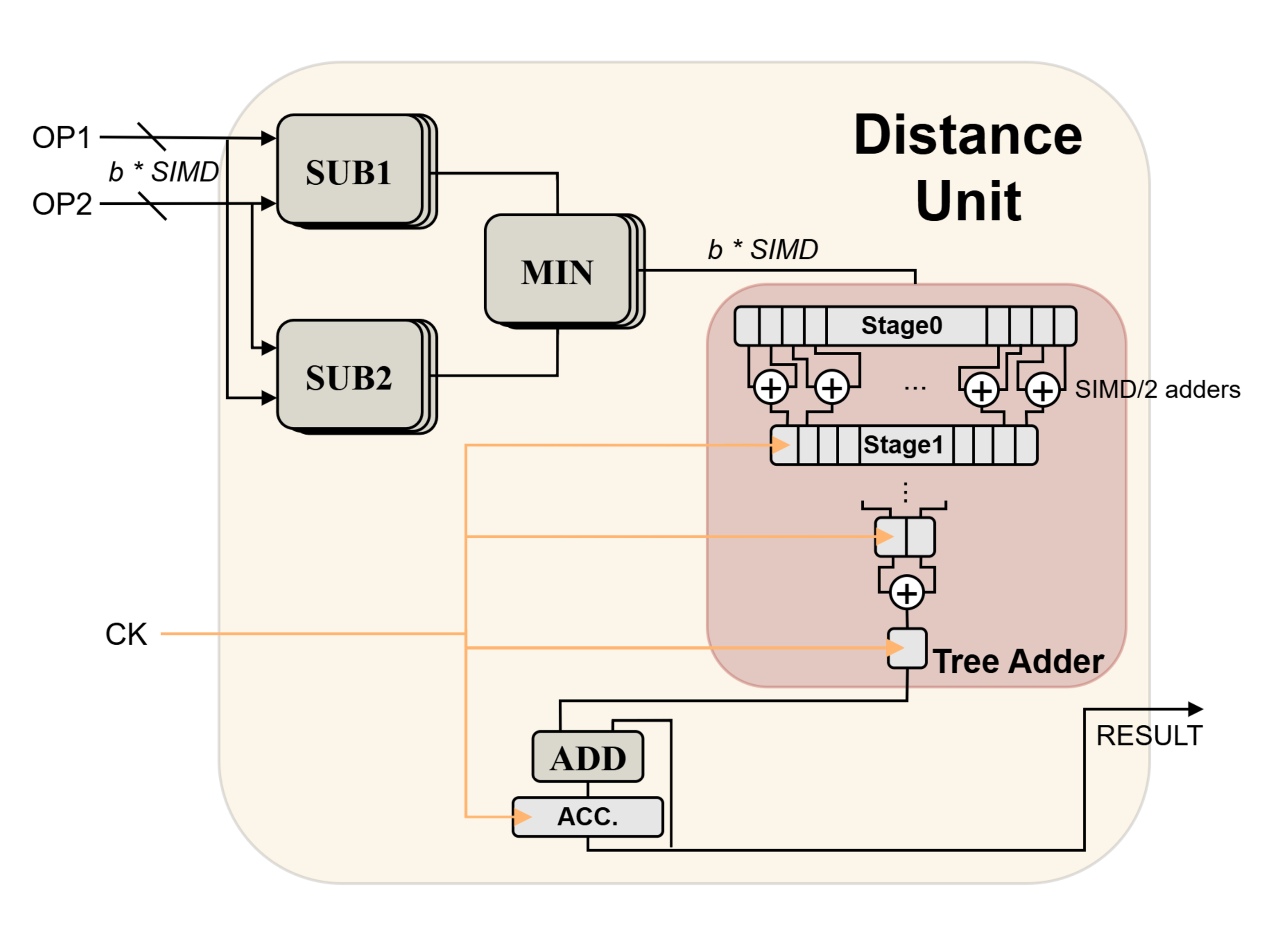}
        \caption{Architecture of the Distance Unit. The design computes $2 \texttt{SIMD}$ differences in parallel, selects the modular distance through a \texttt{min} circuit, and accumulates results using a parallel tree adder.}
        \label{fig: similarity_unit}
    \end{figure}

    \paragraph{Distance metric}  
    In MCR, distance is computed as the angular difference across all components of two HVs, as defined in Eq.~\eqref{eq: similarity}. 
    As in binding, modular reduction is inherently handled by the binary overflow. 
    Accordingly, the unit computes $2 \texttt{SIMD}$ differences in parallel using subtractors, and for each pair of results, a lightweight \texttt{min} circuit selects the correct distance. 
    The resulting per-component distances are then accumulated across \texttt{SIMD} lanes by a parallel tree adder of depth $\lceil \log_2(\texttt{SIMD})\rceil$, as illustrated in Fig.~\ref{fig: similarity_unit}. 
    The total latency of the Distance Unit is thus:
    \[
    L_{\texttt{sim}} = \frac{\texttt{HVDIM}}{\texttt{SIMD}} + \lceil \log_2(\texttt{SIMD})\rceil \;\; \text{cycles}
    \]
    where the first term accounts for the selected hardware parallelism and the second for the depth of the tree adder.

    \paragraph{Superposition}
    This operation requires first mapping each HV component to its equivalent vector, then performing vector addition, and finally normalizing the accumulated result back onto the discrete ring $\mathbb{Z}_r$. As highlighted in Section~\ref{Section: MCR}, the normalization step should only be applied once after all operands have been accumulated, since repeated quantization would cause excessive loss of stored information. For this reason, in MCR-HDCU \textit{we decouple superposition and normalization} into two separate instructions and FUs, enabling intermediate accumulation in fixed-point precision before the final projection back onto $\mathbb{Z}_r$.  
    
    The Superposition Unit is built around two compact LUTs storing precomputed $\cos(\tfrac{2\pi}{r}k)$ and $\sin(\tfrac{2\pi}{r}k)$ values for $k \in [0,r-1]$, that as highlighted in Section~\ref{Section: hardware_friendly}, are small and hardware-friendly. The \texttt{FP} parameter can be specified at synthesis time to balance performance and hardware cost.   
    To maintain the desired precision in cumulative superposition, the unit assumes that the first input HV is already in fixed-point Cartesian form and produces a fixed-point result. The second input, represented in $\mathbb{Z}_r$, is mapped on-the-fly to Cartesian coordinates using cosine and sine LUTs. At each cycle, the unit fetches \texttt{SIMD}/2 real parts and \texttt{SIMD}/2 imaginary parts of the first HV from the dedicated high-bandwidth SPM, while the second HV remains fixed for two cycles. The corresponding cosine values are added to the real components of the first HV, and the sine values to its imaginary components. The results are written back to the SPM at a throughput of \texttt{SIMD}/2 complex components per cycle. The total latency of the operation is therefore:
    \[
    L_{\texttt{superimpose}} = \frac{2  \texttt{HVDIM}}{\texttt{SIMD}} \;\; \text{cycles}
    \]  
    
    \begin{figure}[!t]
        \centering
        \includegraphics[width=0.7\linewidth]{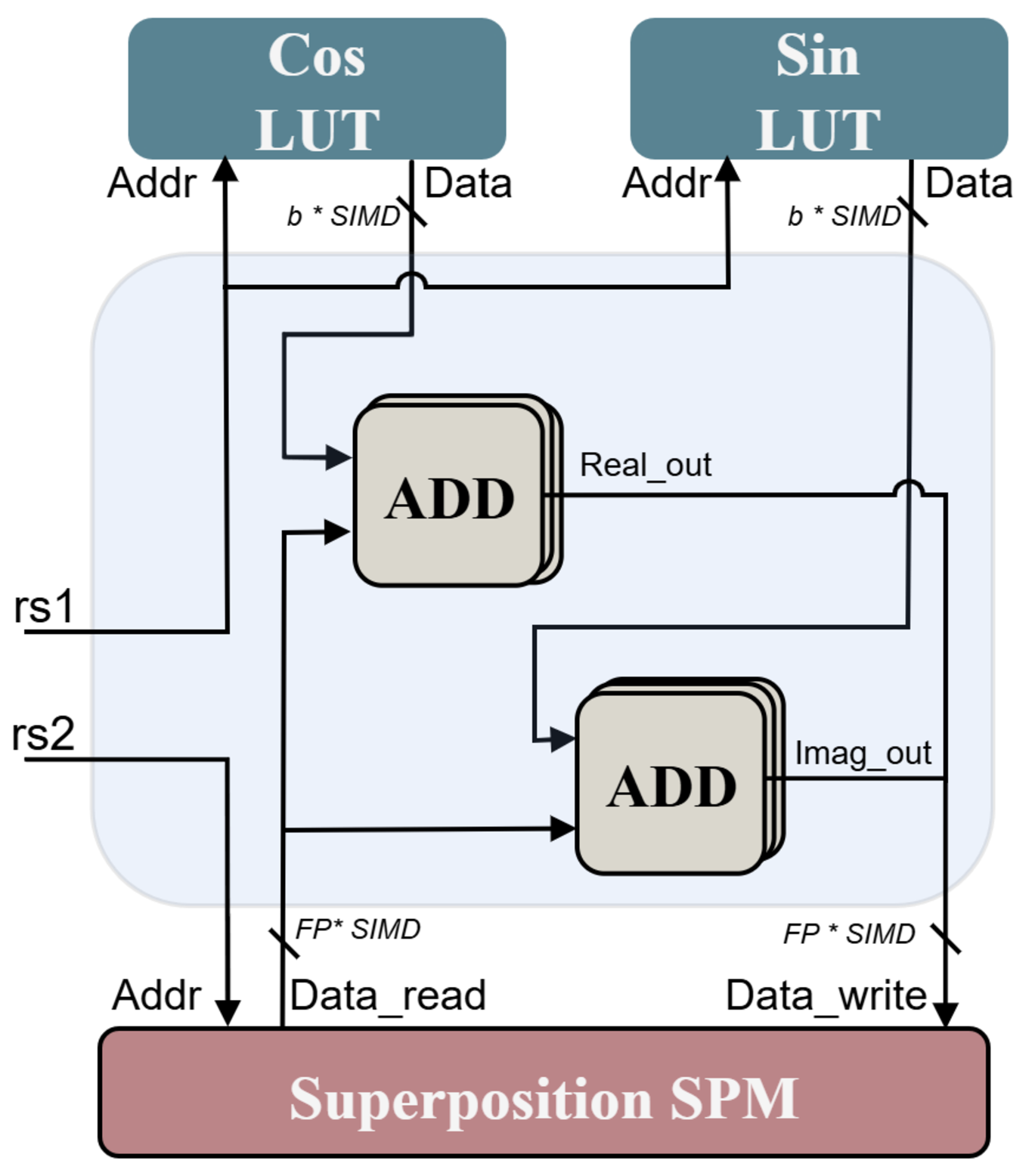}
        \caption{Architecture of the Superposition Unit. Each component is mapped to Cartesian form via LUT-based $\cos/\sin$ values and accumulated in fixed-point precision using dedicated high-bandwidth SPM access.}
        \label{fig: bundling_unit}
    \end{figure}

    \paragraph{Normalization}  
    The Normalization Unit performs the normalization step, projecting the results of superposition that have accumulated in Cartesian coordinates back onto the discrete ring representing $\mathbb{Z}_r$.
    As outlined in Section~\ref{Section: hardware_friendly}, this is achieved through the WTA-based approach, which avoids the costly division and $\operatorname{atan2}$ computations otherwise required by direct normalization.
    
    Given a superimposed component in Cartesian form $(\Re(v_i),\Im(v_i))$, the unit identifies its quadrant from the sign bits of $\Re(v_i)$ and $\Im(v_i)$ and then compares it against the $r/4+1$ directions in that quadrant with coordinates $(\cos \tfrac{2\pi}{r}k,\sin \tfrac{2\pi}{r}k)$ precomputed and stored in the same compact LUTs used by the Superposition Unit.
    Each comparison is implemented as an inner product, requiring two fixed-point multiplications and one addition.
    The accumulated Cartesian HV is stored in a dedicated high-bandwidth SPM that provides simultaneous access to \texttt{SIMD}/2 real and \texttt{SIMD}/2 imaginary parts per cycle. As a result, the total latency of the Normalization Unit is
    \[
    L_{\texttt{norm}} = \frac{2  \texttt{HVDIM}}{\texttt{SIMD}} (r/4+1) \;\; \text{cycles}
    \]

    \paragraph{Permutation}  
    Permutation in HD/VSA produces HVs that are nearly orthogonal to their input HVs.
    Implementing arbitrary rotations in hardware while processing \texttt{SIMD} components in parallel would require complex shuffling logic and intermediate buffers, leading to high area and latency overheads. 
    In this work, we use the insight that even structured permutations (e.g., circular shift or block-cyclic permutation) produce HVs that are approximately orthogonal to input HVs. Using this observation, MCR-HDCU performs permutation at the granularity of whole \texttt{SIMD} blocks. Rather than shifting components within each block, the design cyclically reorders entire blocks by remapping memory addresses. 
    With this design choice, permutation reduces to pure address manipulation in the SPM: read addresses are offset by multiples of \texttt{SIMD} lanes, and the HV is streamed directly in permuted form. This yields $\texttt{HVDIM}/\texttt{SIMD}$ distinct permutations, corresponding to all possible block-level cyclic shifts, while keeping the implementation extremely lightweight. The total latency of the permutation is:
    \[
    L_{\texttt{perm}} = \frac{\texttt{HVDIM}}{\texttt{SIMD}} \;\; \text{cycles}.
    \]
    
     \begin{table}[!t]
        \centering
        \caption{Hardware requirements of MCR-HDCU for different \texttt{SIMD}.
        }
        \resizebox{1\columnwidth}{!}{\begin{tabular}{c|c|c|cccc}\label{tab:simd}
        \textbf{\texttt{SIMD}} & \textbf{Freq.} & \textbf{Module} & \textbf{\#LUTs} & \textbf{\#FFs} & \textbf{\#DSPs} & \textbf{\#BRAMs} \\
        \midrule
        \multirow{2}{*}{8}  & \multirow{2}{*}{150 MHz} & Accelerator & 1707 & 960  & 9  & 0  \\
                            &                          & SPMI  & 726  & 282  & 0  & 16 \\
        \hline
        \multirow{2}{*}{16} & \multirow{2}{*}{125 MHz} & Accelerator & 3139 & 1671 & 17 & 0  \\
                            &                          & SPMI  & 1930 & 486  & 0  & 16 \\
        \hline
        \multirow{2}{*}{32} & \multirow{2}{*}{115 MHz} & Accelerator & 5823 & 3098 & 33 & 0  \\
                            &                          & SPMI  & 3065 & 887  & 0  & 20 \\
        \hline
        \multirow{2}{*}{64} & \multirow{2}{*}{118 MHz} & Accelerator & 13609 & 8564 & 65 & 0  \\
                            &                          & SPMI  & 7193  & 1693 & 0  & 28 \\
        \bottomrule
        \end{tabular}}
    \end{table}
    
    \paragraph{Search}  
    The final operation implemented in MCR-HDCU is \emph{search}, which is essential during inference to identify the closest prototype among a set of HVs. Implementing this purely in software would require issuing the distance instruction $c$ times for a $c$-class problem, limiting the benefits of acceleration. To avoid this overhead, MCR-HDCU integrates a dedicated search instruction that reuses the Distance Unit in a hardware-controlled loop.   
    At runtime, the number of classes is specified as a parameter, enabling the accelerator to iterate over all prototypes in the SPM. The query HV is held constant, while the second operand cycles through each class HV. After each distance computation, the result is compared against the current best match stored in a register, which is updated if necessary. Once the loop completes, the index of the class with the lowest distance is written back to the memory. 
    
     \begin{table*}[!t]
     
        \centering
        \caption{Execution Time in $\mu$s and Energy Consumption in $\mu$J per arithmetic operation in MCR}
        \resizebox{2\columnwidth}{!}{%
        \begin{tabular}{llc|cc|cc|cc|cc|cc}
        & \multirow{2}{*}{\textbf{Realization}} & \multirow{2}{*}{\texttt{HVDIM}} & 
        \multicolumn{2}{c}{\textbf{Binding}} &
        \multicolumn{2}{c}{\textbf{Superposition}} &
        \multicolumn{2}{c}{\textbf{Normalization}} &
        \multicolumn{2}{c}{\textbf{Permutation}} &
        \multicolumn{2}{c}{\textbf{Distance}} \\
        &  &  & \textbf{Time [$\mu$s] } & \textbf{Energy($\mu$J) } & \textbf{Time [$\mu$s]} & \textbf{Energy [$\mu$J]} & \textbf{Time [$\mu$s]} & \textbf{Energy [$\mu$J]} & \textbf{Time [$\mu$s]} & \textbf{Energy [$\mu$s]} & \textbf{Time [$\mu$s]} & \textbf{Energy [$\mu$s]} \\
        \midrule
        & Software & 64 & 5.90 & 0.74 & 13.12 & 1.64 & 76.21 & 9.53 & 11.90 & 1.49 & 4.71 & 0.59 \\
        & Software & 512 & 43.63 & 5.45 & 102.21 & 12.78 & 606.18 & 75.77 & 92.90 & 11.61 & 35.42 & 4.43 \\
        & Software & 2048 & 173.02 & 21.63 & 407.58 & 50.95 & 2411.27 & 301.41 & 380.38 & 47.55 & 140.66 & 17.58 \\
        & \texttt{SIMD} 8 & 64 & 0.15 & 0.02 & 0.36 & 0.04 & 0.65 & 0.07 & 0.15 & 0.01 & 0.24 & 0.02 \\
        & \texttt{SIMD} 8 & 512 & 0.52 & 0.06 & 1.85 & 0.22 & 4.38 & 0.53 & 0.52 & 0.06 & 0.99 & 0.11 \\
        & \texttt{SIMD} 8 & 2048 & 1.80 & 0.22 & 6.97 & 0.93 & 17.18 & 2.32 & 1.80 & 0.22 & 3.55 & 0.44 \\
        & \texttt{SIMD} 16 & 64 & 0.14 & 0.01 & 0.31 & 0.04 & 0.46 & 0.05 & 0.14 & 0.01 & 0.23 & 0.02 \\
        & \texttt{SIMD} 16 & 512 & 0.37 & 0.04 & 1.21 & 0.15 & 2.70 & 0.33 & 0.37 & 0.04 & 0.68 & 0.08 \\
        & \texttt{SIMD} 16 & 2048 & 1.14 & 0.15 & 4.28 & 0.62 & 10.38 & 1.53 & 1.14 & 0.14 & 2.22 & 0.28 \\
        & \texttt{SIMD} 32 & 64 & 0.14 & 0.02 & 0.28 & 0.04 & 0.32 & 0.04 & 0.14 & 0.02 & 0.23 & 0.03 \\
        & \texttt{SIMD} 32 & 512 & 0.26 & 0.04 & 0.77 & 0.12 & 1.54 & 0.25 & 0.26 & 0.04 & 0.47 & 0.06 \\
        & \texttt{SIMD} 32 & 2048 & 0.68 & 0.10 & 2.44 & 0.39 & 5.71 & 0.93 & 0.68 & 0.10 & 1.30 & 0.19 \\
        & \texttt{SIMD} 64 & 64 & 0.13 & 0.02 & 0.24 & 0.04 & 0.23 & 0.04 & 0.13 & 0.02 & 0.20 & 0.03 \\
        & \texttt{SIMD} 64 & 512 & 0.19 & 0.04 & 0.48 & 0.10 & 0.82 & 0.17 & 0.19 & 0.04 & 0.33 & 0.06 \\
        & \texttt{SIMD} 64 & 2048 & 0.39 & 0.08 & 1.30 & 0.27 & 2.86 & 0.59 & 0.39 & 0.08 & 0.74 & 0.15 \\
        \end{tabular}%
        \label{tab:op_merged_time_energy}
        }
     \end{table*}

\subsection{Performance of MCR-HDCU}
\label{section:res:hardware:perf}
    This section presents a comprehensive evaluation of the MCR-HDCU accelerator. 
    We begin by reporting hardware synthesis results, including resource utilization and maximum achievable frequency on FPGA, to characterize the hardware requirements of the proposed accelerator and evaluate the cost of different \texttt{SIMD} configurations.  
    We then analyze performance, evaluating the advantages of the proposed MCR-HDCU on both basic HD/VSA operations and full learning kernels to demonstrate the accelerator's efficiency and flexibility. 
    Finally, we directly compare MCR and BSC by benchmarking the proposed accelerator against the original binary-only HDCU~\cite{HDCU}, denoted here as BSC-HDCU. 
    
    In the experiments below, all benchmarks are implemented using a custom software library that supports two execution modes: a standard (non-accelerated) configuration compiled for the baseline RISC-V instruction set and executed on the Klessydra-T03~\cite{Klessydra-T13} core, and an accelerated configuration compiled with the extended instruction set to exploit the HDCU coprocessor. 

    Execution time is estimated using cycle-accurate Register-Transfer-Level simulations in QuestaSim, by extracting the number of clock cycles and multiplying them by the post-implementation maximum operating frequency of each configuration. Speedup factors are computed as the ratio between baseline and accelerated execution times.  
    For energy consumption, we perform post-implementation gate-level simulations to generate switching activity files (\texttt{.saif}) for each kernel, which are then analyzed using the Vivado power estimator to derive the dynamic power of both the core and the accelerator. 

\subsubsection{Hardware realization}
    To assess the resource requirements and achievable performance of MCR-HDCU, we synthesized the Klessydra-T03 core extended with our MCR-HDCU accelerator on a Xilinx Zynq UltraScale+ ZCU106 (EK-U1-ZCU106-G) device, using Vivado 2023.2.
    Table~\ref{tab:simd} reports the detailed hardware utilization of MCR-HDCU for \texttt{SIMD} widths ranging from 8 to 64, while fixing the modulus $r=16$, fixed-point precision \texttt{FP}=$16$, and using four 2 KB SPMs.
    The reported metrics include LUTs, Flip-Flops (FFs), Digital Signal Processors (DSPs), Block RAMs (BRAMs), as well as the maximum operating frequency achievable on the FPGA.
    
    This analysis highlights how the accelerator can be adapted to different scenarios by balancing hardware cost against performance. For example, increasing the \texttt{SIMD} width from 8 to 64 raises the LUTs count from $2433$ to $20802$, underlining the need to carefully select the degree of parallelism depending on the application. In resource-constrained edge devices primarily targeting inference, a smaller \texttt{SIMD} configuration may be preferable. In contrast, when resources are less constrained and training acceleration is also desired, higher parallelism can be exploited to maximize the throughput.
 
\subsubsection{Impact on the basic HD/VSA arithmetic}
        Table~\ref{tab:op_merged_time_energy} summarizes the execution time and the dynamic energy consumption achieved by MCR-HDCU for each core arithmetic operation.
        All experiments are conducted across a range of HV dimensionalities and hardware parallelism configurations to explore scalability and design trade-offs. Specifically, we evaluate three representative \texttt{SIMD} configurations $\{32, 256, 1024\}$ and three HV dimensionalities $\{64, 512, 2048\}$. 

        As the HV dimensionality increases, the hardware loops inside the accelerator allow efficient processing without re-fetching instructions repeatedly, maximizing the execution speed. On the other hand, increasing \texttt{SIMD} values enables greater parallelism, allowing for the simultaneous processing of more HV components, but at the cost of increased hardware complexity, as reported in Table~\ref{tab:simd}.
        
        For the \textit{binding} operation, MCR-HDCU achieves speedups from $39\times$ ($\texttt{SIMD}=8$, $\texttt{HVDIM}=64$) up to $444\times$ ($\texttt{SIMD}=64$, $\texttt{HVDIM}=2048$), thanks to its lightweight implementation as modular additions with implicit overflow.
        The more complex \textit{superposition} operation benefits from fixed-point Cartesian accumulation using compact cosine/sine LUTs (see Section~\ref{Section: Hardware Design}), and achieves speedups ranging from $36\times$ up to $314\times$. Similarly, the \textit{normalization} operation, implemented via WTA on the precomputed LUTs, provides speedups from $117\times$ to $843\times$, depending on HV dimensionality and \texttt{SIMD} parallelism.
        Much larger gains are observed for the \textit{permutation} operation: by reducing it to a simple hardware-friendly block-wise reordering of memory addresses, MCR-HDCU achieves speedups from $79\times$ to $975\times$.
        Finally, the \textit{distance} metric, that is critical for inference, also benefits strongly from the hardware acceleration, with speedups ranging from $20\times$ to $190\times$, depending on the configuration.
        
        Overall, these results confirm that all core operations of MCR can be executed extremely efficiently in hardware, demonstrating that the additional expressiveness of the modular space does not come at the cost of computational efficiency.
    
        Energy efficiency follows a similar trend. Across all operations, the software implementation on the RISC-V core consumes between $29.5\times$ and $594\times$ more energy than its accelerated counterpart. This reduction highlights the value of specialized hardware, particularly for low-power applications, even when accelerating only individual arithmetic operations.

    \subsubsection{Performance on real data}
\begin{figure}[!t]
        \centering
        \includegraphics[width=0.95\linewidth]{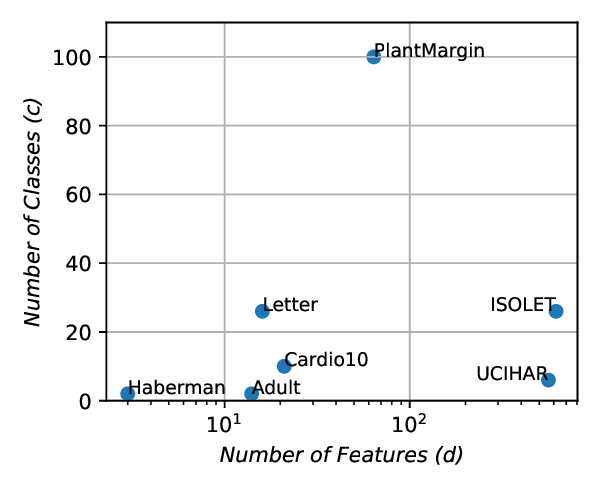}
        \caption{Overview of the real datasets selected for the evaluation, showing their complexity in terms of the number of features and classes. The datasets span a wide range of complexities: \texttt{HabermanSurvival} ($d=3$, $c = 2$), \texttt{Adult} ($d=14$, $c = 2$), \texttt{Letter} ($d=16$, $c = 26$), \texttt{Cardio10} ($d=21$, $c = 10$), \texttt{PlantMargin} ($d=64$, $c = 100$), \texttt{UCIHAR} ($d=561$, $c = 6$), and \texttt{ISOLET} ($d=617$, $c = 26$).}
        \label{fig: datasets}
    \end{figure}

    To demonstrate the flexibility of the proposed accelerator and evaluate its performance in realistic scenarios, we selected seven classification datasets from the collection used in Section~\ref{sec:res:acc}. The selection captures a wide spectrum of problem complexities in terms of both the number of features and classes.
    As shown in Fig.~\ref{fig: datasets}, the selected datasets span from very simple problems such as \texttt{HabermanSurvival} and \texttt{Adult}, with only a few features and two classes, to demanding problems such as \texttt{PlantMargin}, \texttt{UCIHAR}, and \texttt{ISOLET}, which combine hundreds of features with dozens or even hundreds of classes. Intermediate problems such as \texttt{Letter} and \texttt{CARDIO10} represent median cases, containing tens of input features with challenging multi-class classification.
    This variability ensures that the evaluation covers the full range of settings, from lightweight binary problems to high-dimensional, multi-class problems, thereby providing a comprehensive validation of MCR-HDCU. 
    It is important to emphasize that the goal of this analysis is not to assess the classification accuracy, but rather to evaluate the efficiency of the proposed accelerator. Every arithmetic operation implemented in MCR-HDCU was rigorously verified to ensure full consistency with the software model. As a result, the accelerator delivers the exact same accuracy as the reference software implementation reported in Section~\ref{sec:res:acc}.
    
\begin{figure*}
        \centering
        \includegraphics[width=1.9\columnwidth]{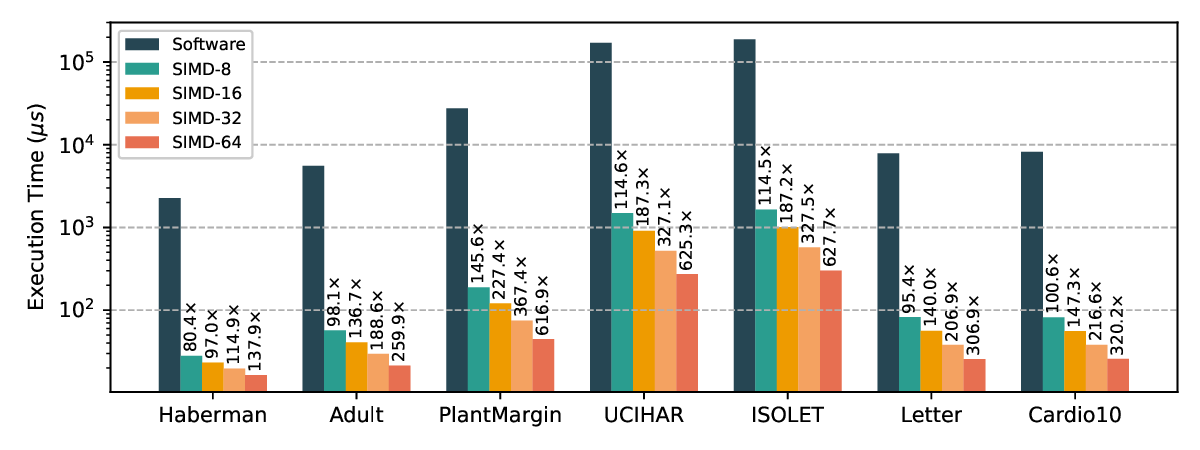}
        \caption{Execution time of $7$ selected datasets on the baseline Klessydra-T03 core compared to the MCR-HDCU accelerator with different \texttt{SIMD} $\in \{8, 16, 32, 64\}$.}
        \label{fig: benchmarks_time}
    \end{figure*}

    For these experiments, we adopted HVs of size $D=1024$ and a modulo $r=16$ (i.e., MCR-4), comparing software execution on the baseline Klessydra-T03 core against accelerated execution with different \texttt{SIMD} configurations \{$8$, $16$, $32$, and $64$\}. 
    Table~\ref{tab:exec_time} reports the obtained execution time for the seven selected datasets, while Fig.~\ref{fig: benchmarks_time} shows the relative speedups. The reported time corresponds to a single inference iteration. In the MCR-HDCU, latency depends only on the number of features and classes and is, therefore, constant across all samples within a dataset.
    
    Across all datasets, the accelerator delivers substantial performance improvements. For the simplest tasks such as \texttt{HabermanSurvival} ($d=3$, $c = 2$), speedups range from $80.4\times$ at $\texttt{SIMD}=8$ to $137.9\times$ at $\texttt{SIMD}=64$, while \texttt{Adult} ($d=14$, $c = 2$) achieves $98.1\times$ to $259.9\times$. Larger datasets yield even higher gains: \texttt{Letter} ($d=16$, $c = 26$) improves from $95.4\times$ to $306.9\times$ while \texttt{Cardio10} ($d=21$, $c = 10$) from $100.6\times$ to $320.2\times$. For the extreme cases with a large number of features and classes, the impact of the hardware acceleration is even more pronounced: \texttt{PlantMargin} ($d=64$, $c = 100$) shows $145.6\times$ to $616.9\times$, \texttt{UCIHAR} ($d=561$, $c = 6$) $114.6\times$ to $625.3\times$, and \texttt{ISOLET} ($d=617$, $c = 26$) achieves the highest gains with $114.5\times$ to $627.7\times$.
    
    Overall, these results confirm that the benefits of hardware acceleration increase with the dataset complexity. In particular, as the number of features increases, the transformation of feature vectors into HVs dominates the runtime and the acceleration of the superposition and binding operations becomes increasingly beneficial. On the other hand, a larger number of classes amplifies the cost of searching for the closest prototype, making the role of the search and distance units the dominant factor.

    \subsubsection{Comparison with BSC}
    The previous sections demonstrated how the arithmetic of MCR can be efficiently implemented in digital hardware devices and how the adoption of a dedicated accelerator can reduce its execution time and energy consumption substantially. At the same time, Section~\ref{sec:res:acc} reported that the higher per-component precision and the modular space allow MCR-4 with $D=64$ components and $r=16$ (4 bits per component) to achieve higher average accuracy (75.49\% across the collection) than BSC with $D=1024$ components (74.35\%), despite requiring $4\times$ less memory. 
    Building on these findings, we conclude the experiments by revisiting the hypothesis from the original MCR study~\cite{MCR} that \textit{BSC is advantageous over MCR in resource-constrained scenarios with strict requirements on execution time and energy efficiency.} To answer this question, we directly compare MCR-4 ($D=64$, $r=16$) against BSC ($D=1024$), when both are accelerated on their corresponding architectures: MCR-HDCU and BSC-HDCU.

    For the sake of fairness, we evaluate four \texttt{SIMD} configurations for each accelerator: $\texttt{SIMD} \in\{8,16,32,64\}$ for MCR-HDCU and $\texttt{SIMD} \in \{32,64,128,256\}$ for BSC-HDCU. These configurations yield an equivalent number of bits processed per clock cycle, since MCR-4 uses $r=16$ ($b=4$ bits per component). Both accelerators are synthesized with four 2 KB SPMs each. Testing \texttt{SIMD} values larger than the dimensionality of HVs  for MCR provides no additional insights, as configurations with $\texttt{SIMD}>\texttt{HVDIM}$ converge to the same execution time and energy consumption as the case when $\texttt{SIMD}=\texttt{HVDIM}$.

    \begin{table}[!t]
        \centering
        \caption{Execution time in $\mu s$ comparing software and \texttt{SIMD} hardware realizations across the selected datasets}
        \resizebox{\columnwidth}{!}{%
        \begin{tabular}{l|rrrrr}
            \hline
            \textbf{Dataset}  & \textbf{Software} & \textbf{SIMD-8} & \textbf{SIMD-16} & \textbf{SIMD-32} & \textbf{SIMD-64} \\
            \hline
            \texttt{HabermanSurvival} & 2258.40  & 28.08  & 23.27  & 19.65  & 16.38 \\
            \texttt{Adult}            & 5565.15  & 56.75  & 40.70  & 29.50  & 21.42 \\
            \texttt{PlantMargin}      & 27495.41 & 188.79 & 120.92 & 74.84  & 44.57 \\
            \texttt{UCIHAR}           & 170264.30 & 1486.03 & 909.21 & 520.56 & 272.30 \\
            \texttt{ISOLET}           & 188503.79 & 1647.03 & 1006.94 & 575.63 & 300.32 \\
            \texttt{Letter}           & 7854.92  & 82.33  & 56.10  & 37.97  & 25.59 \\
            \texttt{Cardio10}         & 8230.95  & 81.83  & 55.88  & 38.00  & 25.70 \\
            \hline
        \end{tabular}} \label{tab:exec_time}
    \end{table}

    \begin{table*}[!t]
        \centering
        \caption{Execution time ($\mu$s) and Energy consumption ($\mu$J) for 
        {BSC ($D=1024$)} and {MCR-4 ($D=64$)}, across datasets and \texttt{SIMD} configurations}
        \label{tab: exe_time_bsc_vs_mcr}
        \resizebox{\textwidth}{!}{%
        \begin{tabular}{lccc|cccccccccccccc}
             & \multirow{2}{*}{\texttt{SIMD}} & \multirow{2}{*}{\#LUTs} & \multirow{2}{*}{\#FFs} 
             & \multicolumn{2}{c}{\texttt{Haberman}} & \multicolumn{2}{c}{\texttt{Adult}} & \multicolumn{2}{c}{\texttt{Letter}} 
             & \multicolumn{2}{c}{\texttt{CARDIO10}} & \multicolumn{2}{c}{\texttt{Plant}} & \multicolumn{2}{c}{\texttt{UCIHAR}} & \multicolumn{2}{c}{\texttt{ISOLET}} \\
             & & & & Time & Energy & Time & Energy & Time & Energy & Time & Energy & Time & Energy & Time & Energy & Time & Energy \\
            \hline
            \multirow{4}{*}{BSC, $D=1024$} 
             & 32  & 1886  & 728  & 6.418 & 0.624 & 21.018 & 2.158 & 27.164 & 2.960 & 31.473 & 3.458 & 87.673 & 10.185 & 747.627 & 87.529 & 824.864 & 97.315 \\
             & 64  & 3255  & 1010 & 4.200 & 0.428 & 11.550 & 1.240 & 14.632 & 1.679 & 16.809 & 1.929 & 45.105 & 5.506 & 377.345 & 46.067 & 416.218 & 50.812 \\
             & 128 & 6547  & 1671 & 3.181 & 0.338 &  7.070 & 0.780 &  8.670 & 1.042 &  9.842 & 1.183 & 24.819 & 3.169 & 200.586 & 25.615 & 221.126 & 28.238 \\
             & 256 & 13094 & 2909 & 2.605 & 0.304 &  4.600 & 0.553 &  5.409 & 0.690 &  6.019 & 0.772 & 13.707 & 1.847 & 103.907 & 13.927 & 114.437 & 15.422 \\
            \hline
            \multirow{4}{*}{MCR-4, $D=64$} 
             & 8   & 2433  & 1242 & 2.173 & 0.235 & 4.447 & 0.507 & 6.140 & 0.743 & 6.320 & 0.771 & 14.887 & 1.920 & 117.720 & 15.304 & 130.360 & 17.077 \\
             & 16  & 5069  & 2157 & 1.976 & 0.219 & 3.736 & 0.437 & 4.824 & 0.603 & 5.112 & 0.639 & 11.800 & 1.569 & 91.400 & 12.156 & 101.000 & 13.433 \\
             & 32  & 8887  & 3985 & 1.750 & 0.224 & 3.217 & 0.428 & 3.883 & 0.563 & 4.283 & 0.621 & 9.917 & 1.527 & 76.233 & 11.740 & 84.033 & 12.941 \\
             & 64  & 20802 & 10257 & 1.687 & 0.270 & 3.026 & 0.499 & 3.470 & 0.607 & 3.939 & 0.693 & 9.122 & 1.688 & 69.661 & 12.818 & 76.652 & 14.181 \\
            \hline
        \end{tabular}}
    \end{table*}
    
    Table~\ref{tab: exe_time_bsc_vs_mcr} reports the results obtained across the previously selected datasets. Despite the intrinsically more complex arithmetic of MCR, the combination of an optimized accelerator design and the reduced number of components enabled by higher per-component precision allows MCR-HDCU to consistently outperform BSC-HDCU in both execution time and energy consumption. On average, across all datasets and \texttt{SIMD} configurations, MCR-HDCU runs kernels {3.08$\times$ faster} and with {$2.68\times$ lower energy consumption} than BSC-HDCU. These advantages are even more pronounced for large datasets such as UCIHAR and ISOLET, where the execution time improves by up to $6.35\times$ while the energy consumption drops by up to $5.72\times$.

    Importantly, the reported improvements do not come at the expense of larger accelerators. For example, when comparing {MCR-4 with $\texttt{SIMD} = 8$} against {BSC with $\texttt{SIMD} = 64$}, an instance where MCR-HDCU is actually smaller in terms of the required resources, MCR-4 still provides improvements: the execution time is reduced by $1.90\times$ to $3.21\times$ while the energy consumption by $1.82\times$ to $3.01\times$, depending on the dataset. 
    
    These results confirm the key insight of this study: although MCR requires more sophisticated arithmetic than BSC, its modular space and higher per-component precision allow much lower HV dimensionality. When paired with efficient hardware support, this leads to significantly faster execution and lower energy consumption. In other words, MCR is not only more accurate per bit but also more efficient when deployed on dedicated hardware.

    \section{Conclusions}
    In this study, we revisited the modular composite representation (MCR) model, providing an extensive analysis of its properties and implementing the first hardware accelerator for this model. 
    Our study demonstrates that the MCR model achieves the best trade-off between information capacity, classification accuracy, and efficiency, positioning it as a compelling alternative to more well-known models such as binary spatter codes \cite{kanerva1995family} and multiply–add–permute \cite{map}.
    
    First, we found that the modular discretized space of MCR offers substantially higher information capacity than binary and integer hypervectors, while approaching the information capacity of complex-valued hypervectors at a fraction of their memory footprint. Second, a large-scale evaluation on $123$ classification datasets confirmed that MCR achieves higher accuracy than binary and low-precision integer hypervectors and can match the performance of binary hypervectors while requiring up to $4\times$ less memory. These results highlight the importance of per-component precision, showing that the larger dynamic range in non-binary modular arithmetic indeed pays off in practice.
    
    Next, we designed MCR-HDCU -- the first hardware accelerator for MCR, demonstrating that its arithmetic maps naturally to digital logic, with modular operations realized implicitly through binary overflow and trigonometric mappings efficiently implemented using compact lookup tables.
    Experimental results on basic operations and seven classification datasets show up to three orders-of-magnitude speedups and substantial energy savings compared to software execution. When compared to binary hypervectors at iso-accuracy and accelerated with the original HDCU, MCR-HDCU achieves on average $2.7\times$ faster execution and $3.9\times$ lower energy consumption.
    
    Overall, our findings refute the intuition that the additional complexity of MCR outweighs its benefits.
    On the contrary, it provides a hardware-friendly alternative that delivers superior information capacity, higher classification accuracy, lower memory footprint, and great efficiency when implemented by specialized hardware.

\bibliographystyle{IEEEtran}
\bibliography{bibliography.bib}

@IEEEtranBSTCTL{IEEEexample:BSTcontrol,
  CTLuse_forced_etal       = "yes",
  CTLmax_names_forced_etal = "3",
  CTLnames_show_etal       = "3" 
}

@article{hersche2021near,
  title={Near-channel classifier: Symbiotic communication and classification in high-dimensional space},
  author={Hersche, Michael and Lippuner, Stefan and Korb, Matthias and Benini, Luca and Rahimi, Abbas},
  journal={Brain Informatics},
  volume={8},
  pages={1--15},
  year={2021},
}

@book{omondi2007residue,
  title={Residue number systems: Theory and implementation},
  author={Omondi, Amos R and Premkumar, A Benjamin},
  volume={2},
  year={2007},
  publisher={World Scientific}
}

@inproceedings{kymn2024binding,
  title={Binding in hippocampal-entorhinal circuits enables compositionality in cognitive maps},
  author={Kymn, Christopher and Mazelet, Sonia and Thomas, Anthony and Kleyko, Denis and Frady, Edward and Sommer, Fritz and Olshausen, Bruno},
  booktitle={Advances in Neural Information Processing Systems (NeurIPS)},
  pages={39128--39157},
  year={2024}
}

@article{fiete2008grid,
  title={What grid cells convey about rat location},
  author={Fiete, Ila R and Burak, Yoram and Brookings, Ted},
  journal={Journal of Neuroscience},
  volume={28},
  number={27},
  pages={6858--6871},
  year={2008},
  publisher={Society for Neuroscience}
}

@inproceedings{kymnoscillator,
  title={Oscillator associative memories facilitate high-capacity, compositional inference},
  author={Kymn, Christopher and Bybee, Connor and Yun, Zeyu and Kleyko, Denis and Olshausen, Bruno A},
  booktitle={New Frontiers in Associative Memories},
  year      = {2025},
}

@article{HundredsClassifiers2014,
  author    = {M. Fernandez-Delgado and E. Cernadas and S. Barro and D. Amorim},
  title     = {Do We Need Hundreds of Classifiers to Solve Real World Classification Problems?},
  journal   = {Journal of Machine Learning Research},
  volume = {15},
  year      = {2014},
  pages = {3133--3181}
}

@misc{Dua2019 ,
    author = {Dua, Dheeru and Graff, Casey},
    year = {2019},
    title = {{UCI} Machine Learning Repository},
    url = {http://archive.ics.uci.edu/ml},
    institution = {University of California, Irvine, School of Information and Computer Sciences},
}

@InProceedings{GaylerJackendoff2003,
	author = {R. W. Gayler},
	title = {Vector Symbolic Architectures Answer {Jackendoff's} Challenges for Cognitive Neuroscience},
	booktitle = {Joint International Conference on Cognitive Science (ICCS/ASCS)},
	year = {2003},
	pages = {133--138},
}

@article{Klessydra-T13,
  title={{Klessydra-T}: Designing Vector Coprocessors for Multithreaded Edge-Computing Cores},
  author={Cheikh, Abdallah and Sordillo, Stefano and Mastrandrea, Antonio and Menichelli, Francesco and Scotti, Giuseppe and Olivieri, Mauro},
  journal={IEEE Micro},
  volume={41},
  number={2},
  pages={64--71},
  year={2021},
}

@article{2021cellular,
  title={Cellular automata can reduce memory requirements of collective-state computing},
  author={Kleyko, Denis and Frady, Edward Paxon and Sommer, Friedrich T},
  journal={IEEE Transactions on Neural Networks and Learning Systems},
  volume={33},
  number={6},
  pages={2701--2713},
  year={2022},
}

@article{kleyko_pieee,
  title={Vector symbolic architectures as a computing framework for emerging hardware},
  author={Kleyko, Denis and Davies, Mike and Frady, Edward Paxon and Kanerva, Pentti and Kent, Spencer J and Olshausen, Bruno A and Osipov, Evgeny and Rabaey, Jan M and Rachkovskij, Dmitri A and Rahimi, Abbas and  F. T. Sommer},
  journal={Proceedings of the IEEE},
  volume={110},
  number={10},
  pages={1538--1571},
  year={2022},
}

@Article{kleyko_review_I,
	title = {A Survey on Hyperdimensional Computing aka Vector Symbolic Architectures, {Part I}: Models and Data Transformations},
	author = {D. Kleyko and D. A. Rachkovskij and  E. Osipov and A. Rahimi},
	journal = {ACM Computing Surveys},
	year = {2022},
	volume = {55},
	number = {6},
	pages = {1--40}
}

@Article{kleyko_review_II,
	title = {A Survey on Hyperdimensional Computing aka Vector Symbolic Architectures, {Part II}: Applications, Cognitive Models, and Challenges},
	author = {D. Kleyko and D. A. Rachkovskij and  E. Osipov and A. Rahimi},
	journal = {ACM Computing Surveys},
	year = {2023},
	volume = {55},
	number = {9},
	pages = {1--52}
}

@article{RW_Benini,
  title={Hardware optimizations of dense binary hyperdimensional computing: Rematerialization of hypervectors, binarized bundling, and combinational associative memory},
  author={Schmuck, Manuel and Benini, Luca and Rahimi, Abbas},
  journal={ACM Journal on Emerging Technologies in Computing Systems},
  volume={15},
  number={4},
  pages={1--25},
  year={2019},
}

@inproceedings{RegHD,
  title={{RegHD}: Robust and efficient regression in hyper-dimensional learning system},
  author={Hernandez-Cano, Alejandro and Zhuo, Cheng and Yin, Xunzhao and Imani, Mohsen},
  booktitle={ACM/IEEE Design Automation Conference (DAC)},
  pages={7--12},
  year={2021},
}

@INPROCEEDINGS{HDCluster,
  author={Imani, Mohsen and Kim, Yeseong and Worley, Thomas and Gupta, Saransh and Rosing, Tajana},
    booktitle={Design, Automation \& Test in Europe Conference \& Exhibition (DATE)},
  title={{HDCluster}: An Accurate Clustering Using Brain-Inspired High-Dimensional Computing}, 
  year={2019},
  pages={1591-1594},
}

@Article{Rachkovskiy_2005,
	title = {Sparse Binary Distributed Encoding of Scalars},
	author = {D. A. Rachkovskij  and S. V. Slipchenko and E. M. Kussul  and T. N. Baidyk},
	journal = {Journal of Automation and Information Sciences},
	year    = {2005},
	volume  = {37},
	number  = {6},
	pages   = {12--23}
}

@inproceedings{penz1987closeness,
  title={The closeness code: An input integer to binary vector transformation suitable for neural network algorithms},
  author={Penz, P. A.},
  booktitle={IEEE First Annual International Conference on Neural Networks (ICNN)},
  pages={515--522},
  year={1987}
}

@inproceedings{ni2022hdpg,
  title={{HDPG}: Hyperdimensional policy-based reinforcement learning for continuous control},
  author={Ni, Yang and Issa, Mariam and Abraham, Danny and Imani, Mahdi and Yin, Xunzhao and Imani, Mohsen},
  booktitle={ACM/IEEE Design Automation Conference (DAC)},
  pages={1141--1146},
  year={2022}
}

@article{bees2015imitation,
  title={Imitation of honey bees’ concept learning processes using vector symbolic architectures},
  author={Kleyko, Denis and Osipov, Evgeny and Gayler, Ross W and Khan, Asad I and Dyer, Adrian G},
  journal={Biologically Inspired Cognitive Architectures},
  volume={14},
  pages={57--72},
  year={2015},
}

@article{schlegel2022comparison,
  title={A comparison of vector symbolic architectures},
  author={Schlegel, Kenny and Neubert, Peer and Protzel, Peter},
  journal={Artificial Intelligence Review},
  volume={55},
  number={6},
  pages={4523--4555},
  year={2022},
}

@article{HDCU,
title={Configurable Hardware Acceleration for Hyperdimensional Computing Extension on {RISC-V}},
journal={TechRxiv},
author={Martino, Rocco and Angioli, Marco and Rosato, Antonello and Barbirotta, Marcello and Cheikh, Abdallah and Olivieri, Mauro},
year={2024},
}

@incollection{map,
	author          = {R. W. Gayler},
	title           = {Multiplicative Binding, Representation Operators \& Analogy},
	pages           = {1--4},
	booktitle       = {Advances in Analogy Research: Integration of Theory and Data from the Cognitive, Computational, and Neural Sciences},
	year            = 1998,
}

@article{Kanerva2009,
	author = {Pentti Kanerva},
	journal = {Cognitive Computation},
	number = {2},
	pages = {139--159},
	title = {Hyperdimensional Computing: An Introduction to Computing in Distributed Representation with High-Dimensional Random Vectors},
	volume = {1},
	year = {2009}
}

@article{verges2025classification,
  title={Classification using hyperdimensional computing: A review with comparative analysis},
  author={Verg{\'e}s, Pere and Heddes, Mike and Nunes, Igor and Kleyko, Denis and Givargis, Tony and Nicolau, Alexandru},
  journal={Artificial Intelligence Review},
  volume={58},
  number={6},
  pages={1-41},
  year={2025},
}

@Article{FradyCapacity2018,
	title = {A Theory of Sequence Indexing and Working Memory in Recurrent Neural Networks},
	author={Frady, E Paxon and Kleyko, Denis and Sommer, Friedrich T},
	journal = {Neural Computation},
	year = {2018},
	volume = {30},
	number = {6},
	pages = {1449--1513}
}

@article{kleyko2023efficient,
  title={Efficient decoding of compositional structure in holistic representations},
  author={Kleyko, Denis and Bybee, Connor and Huang, Ping-Chen and Kymn, Christopher J and Olshausen, Bruno A and Frady, E Paxon and Sommer, Friedrich T},
  journal={Neural Computation},
  volume={35},
  number={7},
  pages={1159--1186},
  year={2023},
  publisher={MIT Press One Rogers Street, Cambridge, MA 02142-1209, USA journals-info~…}
}

@article{torchhd,
  author  = {Mike Heddes and Igor Nunes and Pere Vergés and Denis Kleyko and Danny Abraham and Tony Givargis and Alexandru Nicolau and Alexander Veidenbaum},
  title   = {Torchhd: An Open Source {P}ython Library to Support Research on Hyperdimensional Computing and Vector Symbolic Architectures},
  journal = {Journal of Machine Learning Research},
  year    = {2023},
  volume  = {24},
  number  = {255},
  pages   = {1--10},
}

@article{angioli2025hd,
  title={{HD-CB}: The First Exploration of Hyperdimensional Computing for Contextual Bandits Problems},
  author={Angioli, Marco and Rosato, Antonello and Barbirotta, Marcello and Martino, Rocco and Menichelli, Francesco and Olivieri, Mauro},
  journal={arXiv:2501.16863},
  year={2025}
}

@article{MCR,
  title={Modular composite representation},
  author={Snaider, Javier and Franklin, Stan},
  journal={Cognitive Computation},
  volume={6},
  number={3},
  pages={510--527},
  year={2014},
}

@INPROCEEDINGS{9892030,
  author={Bent, Graham and Simpkin, Chris and Li, Yuhua and Preece, Alun},
  booktitle = {International Joint Conference on Neural Networks (IJCNN)}, 
  title={Hyperdimensional Computing Using Time-To-Spike Neuromorphic Circuits}, 
  year={2022},
  volume={},
  number={},
  pages={1-8},
}

@inproceedings{imani2017exploring,
  title={Exploring hyperdimensional associative memory},
  author={Imani, Mohsen and Rahimi, Abbas and Kong, Deqian and Rosing, Tajana and Rabaey, Jan M},
  booktitle={IEEE International Symposium on High Performance Computer Architecture (HPCA)},
  pages={445--456},
  year={2017},
}

@article{rahimi2017high,
  title={High-dimensional computing as a nanoscalable paradigm},
  author={Rahimi, Abbas and Datta, Sohum and Kleyko, Denis and Frady, Edward Paxon and Olshausen, Bruno A and Kanerva, Pentti and Rabaey, Jan M},
  journal={IEEE Transactions on Circuits and Systems I: Regular Papers},
  volume={64},
  number={9},
  pages={2508--2521},
  year={2017},
}

@InProceedings{kanerva1995family,
  author={Kanerva, Pentti},
	title = {A Family of Binary Spatter Codes},
	booktitle = {International Conference on Artificial Neural Networks (ICANN)},
  pages={517--522},
  year={1995}
}

@book{PlateHolographic2003,
  author={Plate, Tony A},
title={Holographic Reduced Representations: Distributed Representation for Cognitive Structures},
publisher={Stanford: Center for the Study of Language and Information (CSLI)},
year={2003},
}

@inproceedings{frady2022computing,
  title={Computing on functions using randomized vector representations (in brief)},
  author={Frady, E Paxon and Kleyko, Denis and Kymn, Christopher J and Olshausen, Bruno A and Sommer, Friedrich T},
  booktitle={Neuro-Inspired Computational Elements Conference (NICE)},
  pages={115--122},
  year={2022}
}

@inproceedings{bandaragoda2019trajectory,
  title={Trajectory clustering of road traffic in urban environments using incremental machine learning in combination with hyperdimensional computing},
  author={Bandaragoda, Tharindu and De Silva, Daswin and Kleyko, Denis and Osipov, Evgeny and Wiklund, Urban and Alahakoon, Damminda},
  booktitle={IEEE Intelligent Transportation Systems Conference (ITSC)},
  pages={1664--1670},
  year={2019},
}

@ARTICLE{Eggimann2021,
  author={Eggimann, Manuel and Rahimi, Abbas and Benini, Luca},
  journal={IEEE Transactions on Circuits and Systems I: Regular Papers}, 
  title={A 5 uW Standard Cell Memory-Based Configurable Hyperdimensional Computing Accelerator for Always-on Smart Sensing}, 
  year={2021},
  volume={68},
  number={10},
  pages={4116-4128},
  doi={10.1109/TCSI.2021.3100266}}

@article{karunaratne2020memory,
  title={In-memory hyperdimensional computing},
  author={Karunaratne, Geethan and Le Gallo, Manuel and Cherubini, Giovanni and Benini, Luca and Rahimi, Abbas and Sebastian, Abu},
  journal={Nature Electronics},
  volume={3},
  number={6},
  pages={327--337},
  year={2020},
  publisher={Nature Publishing Group UK London}
}

@article{nova2014review,
  title={A review of learning vector quantization classifiers},
  author={Nova, David and Est{\'e}vez, Pablo A},
  journal={Neural Computing and Applications},
  volume={25},
  number={3},
  pages={511--524},
  year={2014},
  publisher={Springer}
}

@inproceedings{tinyhd,
  title={{tiny-HD}: Ultra-efficient hyperdimensional computing engine for {IoT} applications},
  author={Khaleghi, Behnam and Xu, Hanyang and Morris, Justin and Rosing, Tajana},
  booktitle={Design, Automation \& Test in Europe Conference \& Exhibition (DATE)},
  pages={408--413},
  year={2021},
}

@article{FixedHD,
  title={Domain-Specific Hyperdimensional {RISC-V} Processor for {Edge-AI} Training},
  author={Wasif, Sandy A and Wael, Miran and Genssler, Paul R and Azab, Eman and Mashaly, Maggie and Abd El Ghany, Mohamed A and Amrouch, Hussam},
  journal={IEEE Transactions on Circuits and Systems I: Regular Papers},
  year={2025},
}

@ARTICLE{VLNPD,
  author={Angioli, Marco and Barbirotta, Marcello and Cheikh, Abdallah and Mastrandrea, Antonio and Menichelli, Francesco and Jamili, Saeid and Olivieri, Mauro},
  journal={IEEE Transactions on Computers}, 
  title={Design, Implementation and Evaluation of a New Variable Latency Integer Division Scheme}, 
  year={2024},
  volume={73},
  number={7},
  pages={1767-1779},
}

@article{frady2019robust,
  title={Robust computation with rhythmic spike patterns},
  author={Frady, E Paxon and Sommer, Friedrich T},
  journal={Proceedings of the National Academy of Sciences},
  volume={116},
  number={36},
  pages={18050--18059},
  year={2019},
  publisher={National Academy of Sciences}
}

@article{orchard2024efficient,
  title={Efficient hyperdimensional computing with spiking phasors},
  author={Orchard, Jeff and Furlong, P Michael and Simone, Kathryn},
  journal={Neural Computation},
  volume={36},
  number={9},
  pages={1886--1911},
  year={2024},
}

@inproceedings{DiaoGLVQHD2021,
	title = {Generalized Learning Vector Quantization for Classification in Randomized Neural Networks and Hyperdimensional Computing},
	author = {C. Diao and D. Kleyko and J. M. Rabaey and B. A. Olshausen},
	booktitle = {International Joint Conference on Neural Networks (IJCNN)},
	year = {2021},
	pages = {1--9}
}

@article{kleyko2021density,
  title={Density encoding enables resource-efficient randomly connected neural networks},
  author={Kleyko, Denis and Kheffache, Mansour and Frady, E Paxon and Wiklund, Urban and Osipov, Evgeny},
  journal={IEEE Transactions on Neural Networks and Learning Systems},
  volume={32},
  number={8},
  pages={3777--3783},
  year={2021},
}

@ARTICLE{11037669,
  author={Sumanasena, Vidura and de Silva, Daswin and Osipov, Evgeny and Rachkovskij, Dmitri A. and Gayler, Ross W.},
  journal={IEEE Access}, 
  title={Implementing Holographic Reduced Representations for Spiking Neural Networks}, 
  year={2025},
  volume={13},
  pages={116606-116620},
  doi={10.1109/ACCESS.2025.3580582}}

@inproceedings{yu2022understanding,
  title={Understanding hyperdimensional computing for parallel single-pass learning},
  author={Yu, Tao and Zhang, Yichi and Zhang, Zhiru and De Sa, Christopher M},
  booktitle={Advances in Neural Information Processing Systems (NeurIPS)},
  pages={1157--1169},
  year={2022}
}

@inproceedings{kohonen1990improved,
  title={Improved versions of learning vector quantization},
  author={Kohonen, Teuvo},
  	booktitle = {International Joint Conference on Neural Networks (IJCNN)},
  pages={545--550},
  year={1990},
}

@InProceedings{kleyko_2015,
  author={Kleyko, Denis and Osipov, Evgeny and Papakonstantinou, Nikolaos and Vyatkin, Valeriy and Mousavi, Arash},
	title = {Fault Detection in the Hyperspace: Towards Intelligent Automation Systems},
	booktitle = {IEEE International Conference on Industrial Informatics (INDIN)},
  year={2015},
        pages = {1219--1224},
}

@Article{industrial2018,
	title = {Hyperdimensional Computing in Industrial Systems: The Use-Case of Distributed Fault Isolation in a Power Plant},
     author={Kleyko, Denis and Osipov, Evgeny and Papakonstantinou, Nikolaos and Vyatkin, Valeriy},
	journal = {IEEE Access},
	year = {2018},
	volume = {6},
	pages = {30766--30777}
}

@inproceedings{rahimi2016robust,
  title={A robust and energy-efficient classifier using brain-inspired hyperdimensional computing},
  author={Rahimi, Abbas and Kanerva, Pentti and Rabaey, Jan M},
  booktitle={IEEE/ACM International Symposium on Low Power Electronics and Design (ISLPED)},
  pages={64--69},
  year={2016}
}

@article{kymn2025computing,
  title={Computing with residue numbers in high-dimensional representation},
  author={Kymn, Christopher J and Kleyko, Denis and Frady, E Paxon and Bybee, Connor and Kanerva, Pentti and Sommer, Friedrich T and Olshausen, Bruno A},
  journal={Neural Computation},
  volume={37},
  number={1},
  pages={1--37},
  year={2024},
}

\end{document}